# Optimizing Bus Travel: A Novel Approach to Feature Mining with P-KMEANS and P-LDA Algorithms

Hongjie Liu[a,b], Haotian Shi[c,d,*], Sicheng Fu[c], Tengfei Yuan[e,*], Xinhuan Zhang[f], Hongzhe Xu[a], Bin Ran[c]

[a]Xi'an Jiaotong University School of Computer Science and Technology Xi'an 710049, China
[b]Guangdong Planning and Designing Institute of Telecommunications Co., Ltd, Guangzhou 510630, China
[c]Civil and Environmental Engineering, University of Wisconsin-Madison, Madison 53706, US
[d]College of Transportation Engineering, Tongji University, Shanghai, 201804, China
[e]SILC Business School, Shanghai University, Shanghai, 201800, China
[f]College of Engineering, Zhejiang Normal University, Jinhua 321004, China

**Abstract**

Customizing services for bus travel can bolster its attractiveness, optimize usage, alleviate traffic congestion, and diminish carbon emissions. This potential is realized by harnessing recent advancements in positioning communication facilities, the Internet of Things, and artificial intelligence for feature mining in public transportation. However, the inherent complexities of disorganized and unstructured public transportation data introduce substantial challenges to travel feature extraction. This study presents a bus travel feature extraction method rooted in Point of Interest (POI) data, employing enhanced P-KMENAS and P-LDA algorithms to overcome these limitations. While the KMEANS algorithm adeptly segments passenger travel paths into distinct clusters, its outcomes can be influenced by the initial K value. On the other hand, Latent Dirichlet Allocation (LDA) excels at feature identification and probabilistic interpretations yet encounters difficulties with feature intermingling and nuanced sub-feature interactions. Incorporating the POI dimension enhances our understanding of travel behavior, aligning it more closely with passenger attributes and facilitating easier data analysis. By incorporating POI data, our refined P-KMENAS and P-LDA algorithms grant a holistic insight into travel behaviors and attributes, effectively mitigating the limitations above. Consequently, this POI-centric algorithm effectively amalgamates diverse POI attributes, delineates varied travel contexts, and imparts probabilistic metrics to feature properties. Our method successfully mines the diverse aspects of bus travel, such as age, occupation, gender, sports, cost, safety, and personality traits. It effectively calculates relationships between individual travel behaviors and assigns explanatory and evaluative probabilities to POI labels, thereby enhancing bus travel optimization.

## 1. Introduction

As the urgency of addressing global warming and environmental pollution escalates, the world is progressively turning its attention to energy conservation and emission reduction(Cai et al., 2021). Bus travel emerges as an effective solution, offering a suite of benefits including large capacity, eco-friendliness, safety, reliability, and low carbon emissions. By championing public transportation, we can cater to a greater number of commuters on our roads, thus easing individual traffic and alleviating congestion. Such a shift not only supports sustainable urban development but also enhances living conditions. Moreover, bus travel reduces traffic emissions, promotes green travel, and makes optimal use of transportation resources. Given its role in driving economic, social, and urban development, bus travel directly influences urban prosperity and living standards, underlining its indispensability for city dwellers(Berg and Ihlström, 2019).

* Corresponding authors: Haotian Shi (hshi84@wisc.edu); Tengfei Yuan (yuantengfei@shu.edu.cn)

Despite the indispensable role that public transportation plays in contemporary urban development, it is not devoid of significant challenges. These issues manifest in various domains: 1) Illogical placement of bus stations, deficient traffic planning(Ceder et al., 2015), made the inconvenience to residents who have to traverse long distances to reach bus stations. 2) The design of public transportation networks is not reasonable, which frequently neglects key variables such as passenger flow demand and optimal travel distance(Jha et al., 2019). 3) The public facilities lack planning, absence of comprehensive planning for public facilities introduces potential vulnerabilities in their operations, management, and the overall passenger experience. 4) Severe shortage of public resources during peak periods(Gkiotsalitis et al., 2019) and 5) Resources are significantly underutilized in off-peak periods(Gkiotsalitis, Eikenbroek and Cats, 2019). Therefore, while bus travel plays a pivotal role in urban transportation, it also faces substantial obstacles that need to be addressed. Tackling these challenges requires a comprehensive approach to optimize bus travel capacity, enhance service quality, reduce travel times, and improve the overall travel experience.

Efficacious implementation of bus travel optimization strategies is contingent upon accurately describing bus travel behavior and enhancing the quality of bus services in response to fluctuating customer demands(Qi et al., 2018). To this end, it is essential to construct detailed profiles of public transportation passengers that capture hierarchical travel preferences while considering both structured and unstructured travel characteristics, as well as periodic and non-periodic travel behaviors. Supported by these nuanced passengers' profiles, an in-depth analysis of public transport travel preferences can more effectively lay the groundwork for both specialized group travel services and personalized transport optimization strategies. However, current research on optimizing bus travel remains inadequate, with a limited comprehensive examination of the attributes unique to bus travel. This deficiency manifests in suboptimal service quality, station layout, and route planning, leading to low bus occupancy rates. Furthermore, problems such as irregular bus arrivals not only elevate economic costs but also contribute to increased carbon emissions.

To address these problems, an increasing number of emerging technologies are being applied to enhance bus travel optimization and services with the advancement of Internet of Things (IoT) technology. These technologies have paved the way for customized services based on Point of Interest (POI), which have started to be applied to various online social networking and information-sharing platforms. POIs, representing the OTD (O: origin; T: travel chain; D: destination) points of bus trips, have been widely retrieved and recorded. With the integration of clustering mining and feature discovery methods in artificial intelligence, it becomes possible to quickly construct bus travel profiles, analyze travel demand, and provide data support for road traffic planning, urban planning, traffic congestion relief, and personal travel(Chen et al., 2023).

Building on the potential of emerging technologies in enhancing bus travel optimization, it is imperative to delve into data analysis methodologies that can further this objective. Data mining serves as an effective approach for feature analysis and can be categorized into supervised learning, unsupervised learning, and semi-supervised learning. Due to privacy protection concerns, the majority of traffic data is processed using unsupervised techniques, making the application of unsupervised learning more widespread. Various clustering algorithms within unsupervised learning each have their pros and cons; however, from a data-driven perspective, the K-means Clustering Algorithm (KMEANS) (Chabchoub and Fricker, 2014) and Latent Dirichlet Allocation (LDA) (Mantyla et al., 2018) exhibit greater versatility. Consequently, this study will focus on these two algorithms as an entry point for in-depth analysis of passenger's characteristics. The research aims to not only mine and analyse passengers' travel features but also to explore avenues for improving these two algorithms.

KMEANS and LDA approaches offer promising avenues for mining distinctive travel behaviours and features, although they are not without their challenges(Karite et al., 2022). The efficiency of the K-means algorithm can be compromised due to the challenges associated with selecting an optimal value for K, while LDA encounters difficulties in data mining when faced with less-than-ideal topic categorization. However, with the inclusion of POI data, a supplementary element of the map closely related to the purpose of bus travel, these strategies can be optimized and expanded. The POI-based mining method can effectively solve the challenge of selecting the initial K value for KMEANS. Additionally, travel topics can be rapidly delineated by POI classification, and feature pre-classification can be provided for LDA feature mining. Therefore, building upon the KMEANS clustering method and the LDA feature discovery method, the initial parameters of clustering and topic selection are optimized to mine the attributes of bus travel. An improved P-KMEANS clustering algorithm based on POI seeds is proposed. By enhancing the initial K value of the KMEANS algorithm, travel behaviours are identified, bus travel purposes are mined, and travel POI feature seeds are generated. To improve the quality of public transport travel feature mining, a P-LDA feature mining algorithm based on POI seeds is proposed. This algorithm is combined with the results of P-KMEANS clustering and pre-processes different types of POI features to form theme scenes with distinct feature tendencies. This approach improves the LDA algorithm's ability to analyse passengers' intrinsic characteristics, calculates the probability relationship between different individual travel behaviours and travel characteristics, assigns probability values to POI classification label items representing characteristic attributes, and provides probability explanations.

The main contributions of this paper can be summarized into the following four points:

1. The paper introduces a novel approach to the exploration of passenger travel attributes by examining the relationship between passenger travel purpose and POI on a map. This approach enhances our understanding of passenger travel behavior and aids in predicting travel trends.
2. We propose an improved POI-based clustering mining algorithm, P-KMEANS (POI-KMEANS). Using real data sets, we demonstrate that incorporating POI data into the algorithm not only enhances accuracy but also yields more practical results for travel behavior analysis.
3. The paper presents a new POI-based method, P-LDA (POI-Latent Dirichlet Allocation), for analyzing passenger characteristic attributes. Applying P-LDA on a real data set allows us to construct detailed passenger travel profiles, enabling personalized travel services.
4. By integrating cluster mining and feature recognition, we deeply analyze passenger travel characteristics. This analysis provides data-driven insights and theoretical foundations for optimizing travel services, thus contributing to the improvement of urban transport and travel behavior research.

The structure of the paper is as follows: Section 2 provides a brief overview of related work; Section 3 defines the problem and proposes the improved POI-based clustering algorithm and feature mining algorithm; Section 4 demonstrates the cased study of the proposed method; and Section 5 concludes the paper with a summary and outlook.

## 2. Related Works

The clustering method based on GPS is an effective method to study travel characteristics(Roriz et al., 2017). In the pursuit of optimized bus travel, this section offers a literature review exploring existing methodologies. Key focus areas include clustering and feature mining algorithms, particularly the KMEANS and LDA algorithms. While the strengths and limitations of these methods are discussed, the section ultimately emphasizes the need for further development and optimization, particularly in areas such as POI based mining and identification. This section is structured as follows: Section 2.1 reviews existing work on bus travel optimization, Section 2.2 examines feature mining techniques for bus travel data, and Section 2.3 explores methods for mining and identifying POI.

### *2.1. Bus Travel Feature Discovery Algorithm*

Various AI-driven techniques, grouped as supervised, semi-supervised, and unsupervised learning, have found use in feature discovery. However, owing to privacy concerns and specific attributes of bus travel data such as objectivity, de-sensitivity, discreteness, disorder, and unverifiability, unsupervised learning emerges as the preferred choice. This section explores the application of the KMEANS algorithm and the LDA method in this regard, with a particular focus on their role in mining periodic attributes of bus travel data.

#### *2.1.1. KMEANS Cluster Mining*

The KMEANS algorithm is renowned for its effectiveness in personal trajectory mining. Its strengths lie in its high execution efficiency, the ability to handle diverse data types, suitability for spherical clustering, and low sensitivity to individual sample variations. This combination of strengths aligns with the nature of bus travel data, which is characterized by distinct travel patterns. Chabchoub et al.(Chabchoub and Fricker, 2014), Li et al.(Li, 2022), Qi-Fen Yang(Yang et al., 2023), Wang et al.(Wang et al., 2016) and Chu et al.(Chu et al., 2020) applied the KMEANS algorithm in different domains, further emphasizing its versatility. The study's author previously proposed a clustering evaluation algorithm based on the duty cycle(H. J. Liu, 2019 ). However, one significant limitation of KMEANS is its heavy dependence on the 'K' value. This can potentially distort the results of cluster mining and fail to fully capture the relationship between urban hotspots and travel paths. This necessitates the integration of KMEANS with other computational methods to enhance the mining of travel attributes and provide a deeper understanding of urban mobility patterns.

#### *2.1.2. LDA Feature Recognition*

The LDA method, a powerful unsupervised learning topic model, shines in its resilience to noise and outliers, and its efficiency in handling large-scale datasets. The effectiveness of LDA is substantiated by a multitude of studies, such as those by Mika et al.(Mantyla, Claes and Farooq, 2018), Wang et al.(Wang and Hao, 2018), Zeng et al.(Zeng et al., 2021), Yihong Zhang(Zhang et al., 2020) and Li et al.(Li et al., 2021). Despite the LDA model's efficacy in topic differentiation, its performance tends to be suboptimal when distinguishing between closely related topics with minor differences and low inter-topic correlation. Additionally, previous research exploring the correlation between urban POIs and travel attributes often encounters difficulties. These studies typically assign probability values to POI labels to represent characteristic attributes. However, in bus travel scenarios, unsupervised learning struggles with the cross-fusion of unique subject features and the intertwined influence of sub-features, resulting in unstable weight distribution for attribute probabilities. To enhance this process, an advanced classification and mining of topic attributes is recommended, alongside the application of fusion techniques to consolidate feature attributes.

### *2.2. Track Mining and Feature Recognition Based on POI*

POIs, as unique location identifiers, are integral to map navigation and feature extraction, providing essential contextual data for geographical and categorical characteristics. Works by Baral et al.(Baral and Li, 2018), Lim et al.(Lim et al., 2019), Carusotto et al.(Carusotto et al., 2021), Bao et al.(Bao et al., 2012) , and Ahmadi et al.(Ahmadi and Nascimento, 2018) elaborate on their potential in the areas of travel recommendation, analysis, and feature mining. While POI-based mining algorithms are ubiquitously employed in travel recommendation, travel analysis, and feature mining, their use in bus travel feature attribute analysis and mining introduces a set of unique challenges. Among these are insufficient granularity in feature analysis, sub-optimal mining precision, undue influence of subjective experiences on algorithmic performance, low inter-topic correlation, and indistinct topic boundaries. These complexities often impede the effective management of cross-fusion between unique subject features and the mutual influence of sub-features, leading to less than satisfactory mining results. Consequently, there is a pressing need to integrate and refine POI, cluster mining, and feature recognition algorithms to undertake a thorough and efficient analysis of bus travel characteristics.

## 3. Methodology of Bus Travel Feature Mining

Drawing upon the literature review outlined in Section 2, this section embarks on a detailed exploration of the problem statement, bus travel attributes analysis, and the proposed methodology for feature extraction. For the overall structure of this paper, Section 3.1 define and explain the objectives of the paper. Section 3.2 outlines the comprehensive structure of the methodology presented in this paper. Section 3.3 embarks on a thorough qualitative analysis of bus travel characteristics, offering an exhaustive overview. Following this, Section 3.4 unveils a novel strategy of incorporating POI seeds to heighten the efficacy of feature extraction algorithms.

### *3.1. Problem Statement*

#### *3.1.1. Problem Objectives*

In urban settings, the effective optimization of public transportation systems is contingent upon a comprehensive understanding of the travel attributes and behaviors of passengers. This study aims to conduct an in-depth exploration of a spectrum of passenger attributes, collectively represented as $P = [P_a, P_o, P_g, P_h, P_e, P_s, P_p]$. Here, $P_a$ refers to the age of the passenger, $P_o$ to their occupation, $P_g$ to their gender, $P_h$ to their health status, $P_e$ to their economic background, $P_s$ to their safety concerns, and $P_p$ to their personality traits. To clarify the notations used in this paper, Table 1 presents a detailed breakdown of the input and output variables.

By analysing and leveraging these attributes, the study seeks to enhance the efficiency and attractiveness of public transportation systems. Specifically, we aim to unearth the inherent passenger attributes and behaviors of urban commuters. The objectives are outlined as follows:

1. Uncovering travel attributes**:** Focusing on seven commonly observed attributes closely related to passenger travel behaviors, and utilizing them for feature extraction and combinatorial research(Chen, Liu, Lyu, Vlacic, Tang and Liu, 2023).
2. Correlational analysis**:** Investigating the correlation between travel behavior features and Points of Interest (O), to establish a foundational understanding that aids in predicting future travel patterns.

Table 1 Input and Output notation table

| Notation | Descriptions |
|---|---|
| $T$ | the passenger travel trajectories |
| $T_l$ | the trajectory longitude |
| $T_d$ | the trajectory latitude |
| $T_t$ | the trajectory timestamp |
| $O$ | the point of interest |
| $O_l$ | the POI longitude |
| $O_d$ | the POI latitude |
| $O_n$ | the POI name |
| $O_{lb}$ | the POI label category |
| $O_a$ | the POI address |
| $P$ | the passengers |
| $P_a$ | the age of passenger |
| $P_o$ | the occupation of passenger |
| $P_g$ | the gender of passenger |
| $P_h$ | the health of passenger |
| $P_e$ | the economics of passenger |
| $P_s$ | the safety of passenger |
| $P_p$ | the personality of passenger |
| $I$ | the type of POI |
| $I_c$ | the company type POI |
| $I_h$ | the hotel type POI |
| $I_g$ | the government type POI |
| $I_e$ | the education type POI |
| $I_m$ | the medicine type POI |
| $I_t$ | the traffic type POI |
| $I_{ca}$ | the car type POI |
| $I_{me}$ | the media type POI |
| $I_s$ | the service type POI |
| $I_f$ | the finance type POI |
| $I_{sh}$ | the shopping type POI |
| $I_b$ | the beauty type POI |
| $I_{en}$ | the entertainment type POI |
| $I_{sp}$ | the sports type POI |
| $I_{fo}$ | the food type POI |
| $I_{tr}$ | the travel type POI |
| $\theta$ | the weights of POI |
| $\theta_c$ | the company weight of POI |
| $\theta_g$ | the government weight of POI |
| $\theta_e$ | the education weight of POI |
| $\theta_m$ | the medicine weight of POI |
| $\theta_t$ | the traffic weight of POI |
| $\theta_{ca}$ | the car weight of POI |
| $\theta_{me}$ | the media weight of POI |
| $\theta_s$ | the service weight of POI |
| $\theta_f$ | the finance weight of POI |
| $\theta_{sh}$ | the shopping weight of POI |
| $\theta_b$ | the beauty weight of POI |
| $\theta_{en}$ | the entertainment weight of POI |
| $\theta_{sp}$ | the sports weight of POI |
| $\theta_{fo}$ | the food weight of POI |
| $\theta_{tr}$ | the travel weight of POI |

*3.1.2. Mathematical Framework of the Problem*

The above introduced passenger attributes, represented by ($P$), are subject to a wide range of factors, encompassing the trip's purpose and diverse real-world constraints. Within this complex system, Points-of-Interest ($O$) emerge as critical variables. Our goal is to develop a model that deciphers the intricate interplay between P and O, enabling the quantification of feature attributes and the weighting of POIs, denoted as θ. To construct this model, we gather empirical data on travel trajectories (T), which are then integrated into a tangible geographical context. This integration ensures a realistic representation of the environment. The model is further refined by superimposing this data with the geographically mapped coordinates of POIs, thereby enhancing the spatial precision and contextual relevance of our analysis.

The mathematical model we have developed is designed for the inferential analysis of feature attributes in $P$ and $O$, as well as their corresponding weightings ($\theta$). This model processes three primary categories of input data:

1. **Passenger Travel Trajectoires (*T*)**: This is represented as $T = [T_l,\ T_d,\ T_t]$, where:
   - $T_l$: Trajectory longitude, indicating the east-west position of the trajectory on the map.
   - $T_d$: Trajectory latitude, indicating the north-south position of the trajectory on the map.
   - $T_t$: Trajectory timestamp, marking the time at which each point in the trajectory was recorded.
2. **Points-of-Interest Information (*O*)**: This is denoted as $O = [O_l,\ O_d, O_n, O_{lb}, O_a]$, where:
   - $O_l$: POI longitude, representing the east-west position of the POI on the map.
   - $O_d$: POI latitude, representing the north-south position of the POI on the map.
   - $O_n$: POI name, providing a textual descriptor of the POI.
   - $O_{lb}$: POI label category, offering a high-level categorization of the POI (e.g., hotel, government building, school).
   - $O_a$: POI address, detailing the physical location of the POI (e.g., street address, city, state).
3. **Types of Points of Interest ( $I$ )**: These are represented by $I = [I_c, I_h, I_g, I_e, I_m, I_t, I_{ca}, I_{me}, I_s, I_f, I_{sh}, I_b, I_{en}, I_{sp}, I_{fo}, I_{tr}]$, where each element signifies a specific POI type as follows:
   - $I_c$: General category of OI.
   - $I_h$: Hotel type POI.
   - $I_g$: Government type POI.
   - $I_e$: Education type POI.
   - $I_m$: Medicine type POI.
   - $I_t$: Traffic type POI.
   - $I_{ca}$: Car type POI.
   - $I_{me}$: Media type POI.
   - $I_s$: Service type POI.
   - $I_f$: Finance type POI.
   - $I_{sh}$: Shopping type POI.
   - $I_b$: Beauty type POI.
   - $I_{en}$: Entertainment type POI.
   - $I_{sp}$: Sports type POI.
   - $I_{fo}$: Food type POI.
   - $I_{tr}$: Travel type POI.

This comprehensive categorization and detailed input data structure enable the model to perform nuanced inferential mining, thereby extracting meaningful insights from the complex interplay of travel trajectories and points of interest.

*3.1.3. Model Output of the Problem*

The model aims to produce a set of predictive indicators encapsulated by $P'$ and $\theta$. Specifically, the objective is to output inferred travel attributes $P'$ and Points-of-Interest weightings $\theta$.

1. **Inferred Passengers attributes ( $P'$ ):** The attributes are represented by $P' = [P'_a, P'_o, P'_g, P'_h, P'_e, P'_s, P'_p]$ . The concepts correspond to the travel attributes in $P = [P_a, P_o, P_g, P_h, P_e, P_s, P_p]$, albeit with different values.

2. **Points-of-Interest weightings ( $\theta$ )**: The attributes are represented by $\theta = [\theta_c, \theta_g, \theta_e, \theta_m, \theta_t, \theta_{ca}, \theta_{me}, \theta_s, \theta_f, \theta_{sh}, \theta_b, \theta_{en}, \theta_{sp}, \theta_{fo}, \theta_{tr}]$, where each element signifies a specific POI type weight as follows:

   - $\theta_c$: Company weight of POI.
   - $\theta_g$: Government weight of POI.
   - $\theta_e$: Education weight of POI.
   - $\theta_m$: Medicine weight of POI.
   - $\theta_t$: Traffic weight of POI.
   - $\theta_{ca}$: Car weight of POI.
   - $\theta_{me}$: Media weight of POI.
   - $\theta_s$: Service weight of POI.
   - $\theta_f$: Finance weight of POI.
   - $\theta_{sh}$: Shopping weight of POI.
   - $\theta_b$: Beauty weight of POI.
   - $\theta_{en}$: Entertainment weight of POI.
   - $\theta_{sp}$: Sports weight of POI.
   - $\theta_{fo}$: Food weight of POI.
   - $\theta_{tr}$: Travel weight of POI.

By solving this analytical inference problem, we take a pivotal step toward achieving a more optimized and passenger-centric public transportation system. The details of this model's architecture, including the relationship between inputs $T$, $O$, $I$, $P$ and outputs $P'$, $\theta$, are elucidated in Fig. 1.

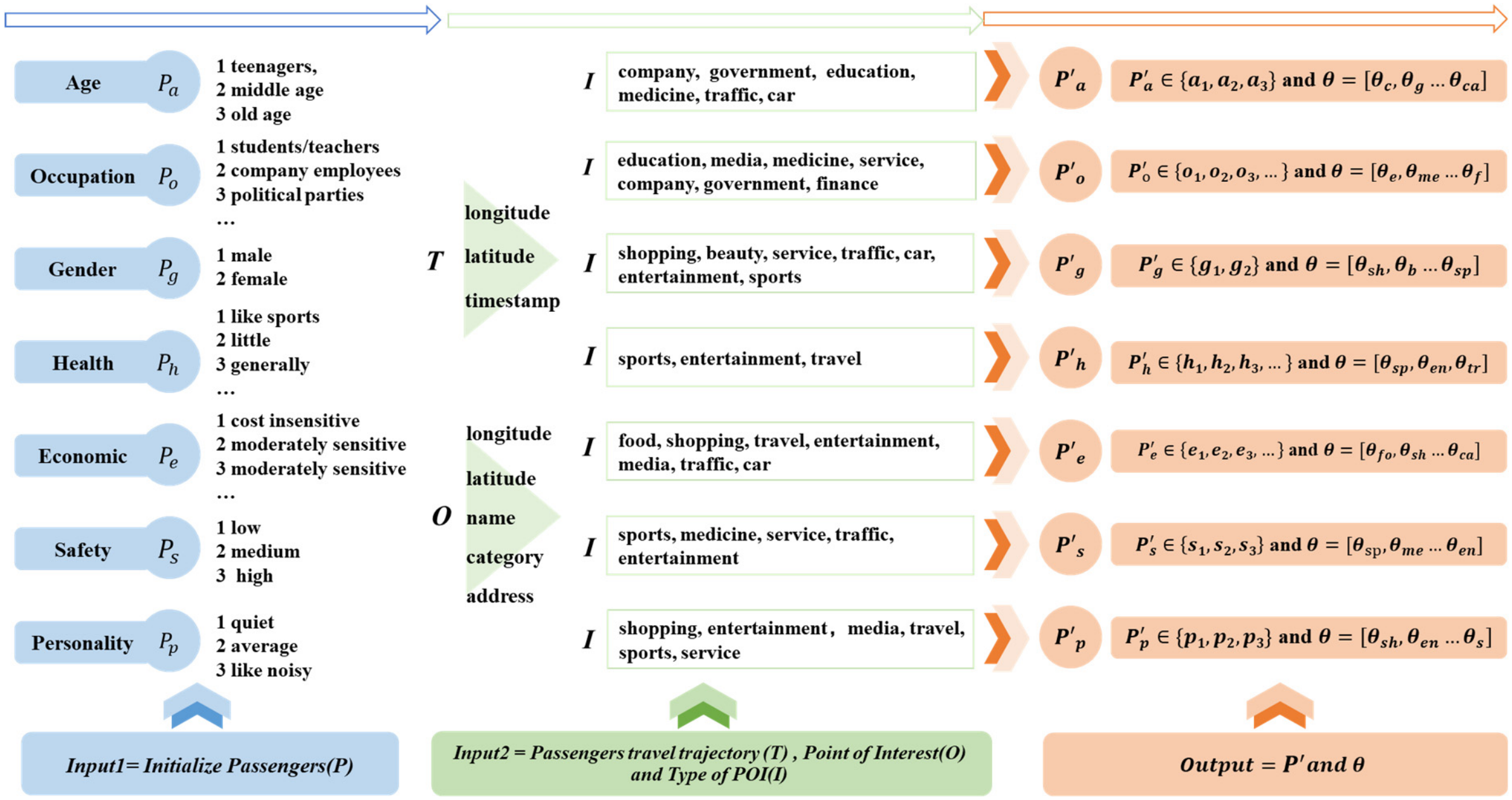

Fig. 1 Mathematical data flow in the model.

Fig. 1 Diagram illustrates the data flow and relationships. In Fig. 1, the model ingests three primary categories of input data: passengers ($P$), Points-of-Interest (POI), and travel trajectories($T$). Through advanced data mining and inferential techniques, the model aims to effectively uncover the inferred feature attributes in $P'$ and quantify the weightings of Points-of-Interest ($\theta$). This multi-faceted approach serves as the foundational framework for our analytical model, designed to provide insightful data that can be instrumental in understanding and predicting passenger travel behavior and preferences.

Through this comprehensive approach that integrates mathematical modeling with real-world data, we aim to make substantive contributions to the optimization of public transportation systems, thereby promoting more sustainable and passenger-friendly travel options.

### *3.2. Methodology Overall Structure*

Based on the problem description defined in Section 3.1, this section introduces a novel methodology rooted in POI trajectory mining and feature detection. This groundbreaking approach leverages a qualitative assessment of travel characteristics, complemented by relevant data mining techniques, as depicted in Fig. 2. The figure is organized into two critical components. The first component (the upper half of Fig. 2) describes the logical flow of the methodology framework. Following the framework, the second component (the lower half of Fig. 2) pertains to the implementation details of the proposed methodology.

The first component of this study is summarized in two main approaches: a qualitative analysis based on passenger travel trajectories and a quantitative analysis based on travel objectives. For the travel trajectories of passengers (bus travel attributes discovery, given in Section 3.3), we integrate with the POI (POI seed model, given in Section 3.4.1) to conduct a qualitative analysis of features closely related to passenger travel behaviors (qualitative analysis of travel on POI seed, given in Section 3.4.2). Furthermore, we employ the POI seed model for a quantitative assessment of passenger travel characteristics, analyzing the impact of these features on travel behaviors (quantitative analysis of bus travel based on POI seed, given in section 3.4.3).

In the second component, we initially classify and identify POIs through a POI capture analysis, which serves as a conduit for constructing the preliminary POI seeds (given in Section 3.4.1). Following this, the POI seeds are integrated into the KMEANS algorithm structure, augmenting both its K value and central value (given in Section 3.4.2). The augmented P-KMEANS algorithm is then employed for the clustering and separation of passenger travel trajectories, thereby generating distinct activity circles for each passenger. In the final phase, both the POI seeds and the resulting life circles (travel radius of the passenger) are integrated into the LDA algorithm to enhance its topic differentiation process (given in Section 3.4.3). By utilizing the enhanced P-LDA model, we can accurately calculate the proportional weight of a passenger’s bus travel characteristics.

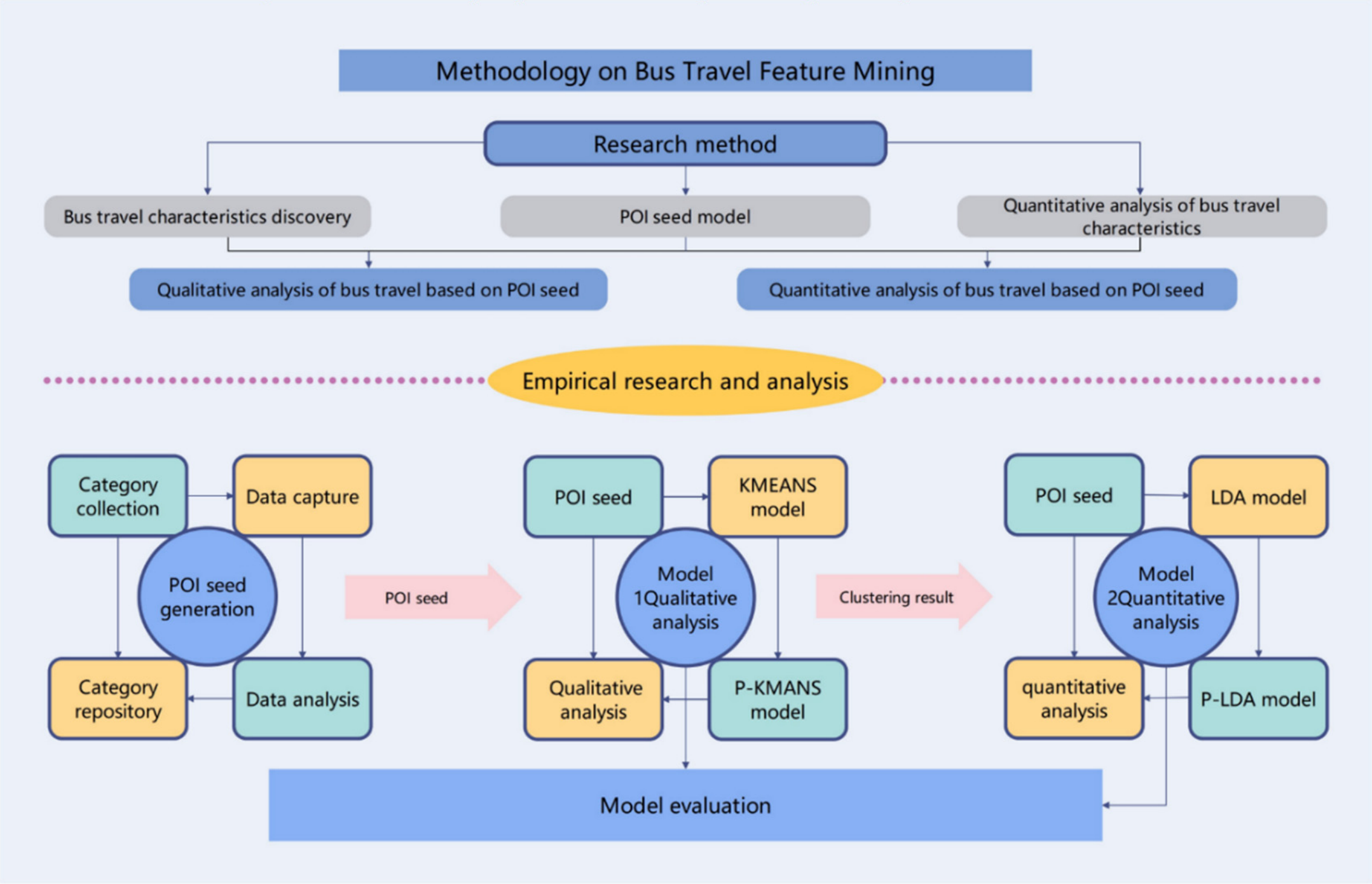

Fig. 2 Overall framework of the proposed bus travel feature mining method.

### *3.3. Attributes of Public Transport Trips*

This section undertakes a detailed study and summation of the defining attributes of bus travel utilizing qualitative analysis. For optimal traffic management, it is crucial to consider both the local and global attributes of urban roads. Additionally, it is essential to comprehend the travel needs and habits of commuters, accurately portray travel profiles based on bus travel dynamics, and organize the travel preferences of various passengers at distinct levels, factoring in both structured and unstructured travel characteristics.

To optimize travel attributes and differentiate between cyclical and non-cyclical travel behaviours, it is indispensable to delve deep into the travel preferences of public transport passengers and base our understanding on comprehensive public transport travel profiles. This approach will significantly aid in laying the groundwork for group travel services and provide pertinent guidance for optimizing public transport based on individual necessities.

Table 2 offers a comprehensive delineation of the travel attributes pertaining to public transportation passengers. It's critical to acknowledge that travel behaviours are intricately intertwined with personal commuting patterns. For instance, passengers frequently availing themselves of direct bus services may demonstrate a higher tolerance for more extended routes. Conversely, those who prioritize punctuality are inclined to pay particular attention to the overall journey duration. In contrast, individuals favouring economical commuting tend to opt for the most cost-efficient modes of travel. These diverse travel behaviours are rooted in individual commuting habits, exhibiting a close relationship with specific travel characteristics(Chen, Liu, Lyu, Vlacic, Tang and Liu, 2023).

Table 2 Attribute description of the transit passengers.

| Bus passengers attributes | | Bus passengers attribute description |
|---|---|---|
| Gender | | 1. Male 2. Female |
| Age | | 1. 11-15 years old 2. 16-19 years old 3. 20-30 years old 4. 31-40 years old 5. 41-50 years old 6. 50-60 years old 7. 60 years old and above |
| Occupation | | 1. Public utility enterprise personnel (including finance) 2. Non-public utility Enterprise personnel 3. Government/participating public institution personnel 4. Education/health/scientific research unit personnel 5. Self-employed/freelance personnel 6. Students 7. Retired 8. Other_____ |
| Possession of a driver's license | | 1. Yes 2. No |
| Possession of a monthly pass | | 1. Yes 2. No |
| Income status | | 1. 1500 yuan and below 2. 1501-3000 yuan 3. 3001-5000 yuan 4. 5001-8000 yuan 5. 8001-15000 yuan 6. More than 15000 yuan |
| Car ownership status | | 1. 0 vehicles 2. 1 vehicle 3. 2 vehicles 4. 3 vehicles and above |
| Bike ownership status | | 1. 0 vehicles 2. 1 vehicle 3. 2 vehicles 4. 3 vehicles and above |
| Health status | | 1. Like sports and quietness 2. Dislike sports, Like quietness 3. Dislike sports, Dislike quiet |
| Bus trip frequency | | 1. 0 times/day 2. 1 time/day 3. 2 times/day 4. >2 times/day |
| Travel time | The time from O to bus station | 1. <5min 2. 5~10min 3. 10~15min 4. 15~20min 5. >20min |
| | Waiting time | 1. 1~3min 2. 3~5min 3. 5~10min 4. 10~15min 5. >15min |
| | Time on the bus | 1. Within 15min 2. 15-30min 3. 30-60min 4. Over 60min |
| | The time from the bus station to D | 1. <5min 2. 5~10min 3. 10~15min 4. 15~20min 5. >20min |
| Purpose of travel | | 1. Going to work 2. Going to school 3. Shopping 4. Picking up children 5. Others |
| Trip distance | Travel distance | 1. <1 *km* 2. 1-3 *km* 3. 3-5 *km* 4. 5-10 *km* 5. >10 *km* |
| | The distance from O to bus station | 1. <50 *m*, 2. 50-100 *m*, 3. 100-200 *m*, 4. 200-500 *m*, 5、>500 *m* |
| | The distance from station to D | 1. <1 *km* 2. 1-3 *km* 3. 3-5 *km* 4. 5-10 *km* 5. >10 *km* |
| | The distance on the bus | 1. <50 *m* 2. 50-100 *m* 3. 100-200 *m* 4. 200-500 *m* 5. >500 *m* |
| Transfer attribute | Transfers number | 1. 0 times 2. 1 times 3. 2 times 4. >2 times |
| | Time | 1. <5min 2. 5~10min 3. 10~15min 4. More than 15min |
| | Distance | 1. <50m 2. 50~100m 3. 100~150m 4. More than 150m |
| | Difficulty level | 1. Easy 2. Moderate 3. Difficult |

### *3.4. Mining Algorithm of Bus Travel Feature Based on POI Seed*

Building on the analyses presented in Section 3.3, we incorporate POI into the frameworks of both the KMEANS and LDA algorithms to enhance the performance of feature extraction procedures in Section 3.4. This innovative step bolsters the variable selection process within these algorithms, with the primary aim being the efficient extraction of distinctive attributes from passenger bus travel data. The overall algorithm framework is depicted in Fig. 3. As shown in Fig. 3, the model is composed of three distinct components: the seed generation algorithm, the P-KMEANS, and the P-LDA. The seed generation algorithm serves as the initial step, capturing POIs and generating the respective seeds. Subsequently, the P-KMEANS is deployed, mining travel life circles predicated on the earlier generated POI seeds. The final component, the P-LDA, then utilizes these travel life circles, along with the POI distribution and weight ratio, to mine the residents' travel attributes and their corresponding weights.

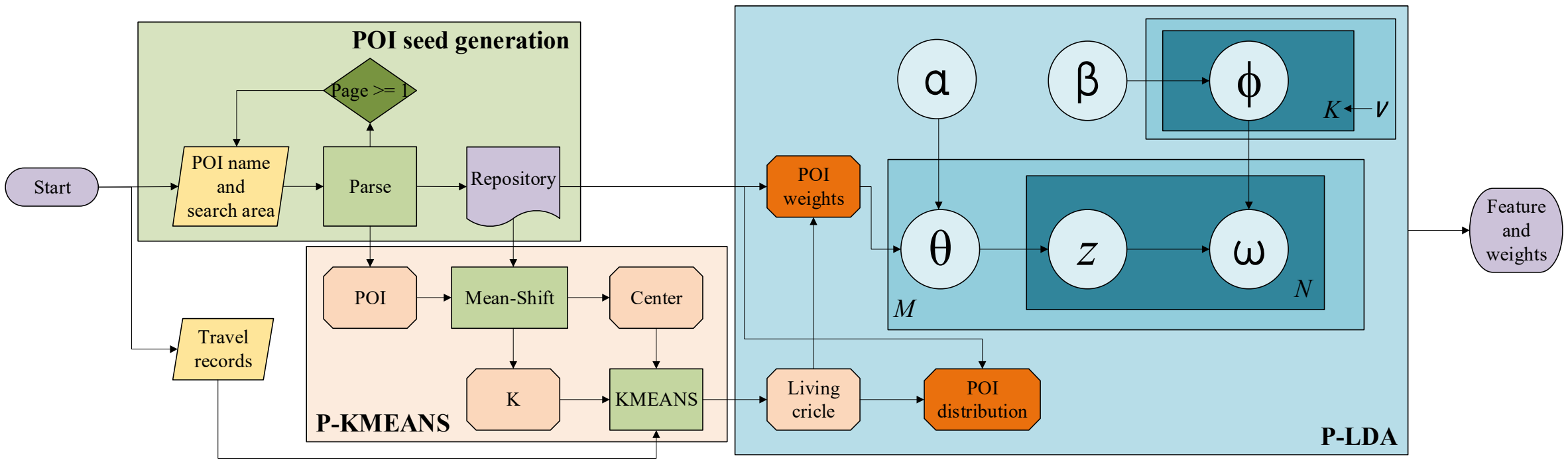

Fig. 3 Mining algorithm for the bus travel features based on POI seed.

In the ensuing subsections, the methodologies for integrating our analysis are further elaborated. Subsection 3.4.1 describes the methodology used for generating POI seeds. Subsection 3.4.2 discusses the process of incorporating POI into the K-Means algorithm to execute an exhaustive cluster analysis. This technique utilizes travel data to formulate a foundational dataset comprising passenger travel trajectories. Subsection 3.4.3 then amplifies the LDA feature mining algorithm applied to the passenger travel trajectory dataset. By incorporating POIs as a means to demarcate interest points, we enhance the algorithm's capability. As a result, this approach facilitates the extraction and generation of feature weights that correspond closely with passenger travel patterns.

#### *3.4.1. POI Seed Generation*

This section describes the process of mining and generating POI seeds. The captured and processed POI data, as described previously, represent key focal points within the current area. Before proceeding with the travel trajectory clustering calculations, these POIs require classification. The selected classification technique involves the Mean-Shift clustering algorithm, which promptly identifies the central location of each interest point group - each representing unique categories of travel purpose - through the mean-shift process. The number of clusters correlates with the number of central hotspots within the area, with the core of the Mean-Shift cluster serving as the initial seed for a POI.

The Mean-Shift algorithm is a hill-climbing algorithm based on kernel density estimation, which can be applied to various tasks such as clustering, image segmentation, and tracking, among others. The process of the Mean-Shift algorithm is outlined as follows:

(a) Choose an unmarked point 'x' in the space to serve as the center of a circle with radius 'h', thereby forming a high-dimensional sphere. Register all points $x_i$ within this sphere.

(b) Compute the current Mean-Shift vector $M_{h,G}(x)$. If $M_{h,G}(x) < \varepsilon$ (less than a pre-specified threshold), the clustering process for 'x' concludes. If not, update 'x' according to M and return to step (a).

(c) Repeat the aforementioned steps until all points have been visited and marked.

The fundamental structure of the algorithm can be articulated as follows. Given 'n' data points 'X' in a 'd'-dimensional space, the basic form of the Mean-Shift vector, M, for any point 'x' in the space is defined as Eq. (1):

$$M_h = \frac{1}{K}\sum_{x_i \in S_k}(x_i - x) \tag{1}$$

where '$S_h$' represents the data point for which the distance from the point within the dataset to 'x' is less than the spherical radius 'h', that is Eq. (2):

$$S_h(x) = \left\{ \mathrm{y} : (y - x_i)_T (y - x_i) < h^2 \right\} \tag{2}$$

The algorithm iteratively updates 'x' such that the centre of the sphere perpetually moves towards the region of maximum data density. In other words, during each iteration, the new centre is defined as the mean position of the points within the sphere. By incorporating the kernel function into the basic Mean-Shift vector, we obtain the Mean-Shift vector as Eq. (3):

$$M_{h,G}(x) = \frac{\sum_{i=1}^{n} x_i g\left(\left\|\frac{x - x_i}{h_i}\right\|^2\right)}{\sum_{i=1}^{n} g\left(\left\|\frac{x - x_i}{h_i}\right\|^2\right)} - x \tag{3}$$

Finally, the coordinates of the new centre, 'x', can be acquired if and only if $M_{h,G}(x) = 0$, such as Eq. (4):

$$x = \frac{\sum_{i=1}^{n} x_i g\left(\left\|\frac{x - x_i}{h_i}\right\|^2\right)}{\sum_{i=1}^{n} g\left(\left\|\frac{x - x_i}{h_i}\right\|^2\right)} \tag{4}$$

### *3.4.2. P-KMEANS Algorithm Optimization Based on POI*

After generating the POI seed, we propose P-KMEANS clustering algorithms to explore and analyze passenger travel patterns based on POI. While prior research has advanced the K-Means algorithm both experimentally and algorithmically, generating noteworthy insights, it often overlooks its enhancement for real-world applications. Traffic data, significant in both research and application, unveils varied perspectives depending on the analytical approach. It's vital to recognize the link between travel choices and purposes, frequently tied to POIs, when refining algorithms and conducting data mining.

In reality, most bus journeys start from a distinct location and are driven by a specific purpose. These locational markers align with particular POIs, each indicating a unique travel path. This data is not only practically valuable but also intuitively insightful. In transportation research, pedestrian travel data mirrors individual preferences, reflecting particular key traits. Thus, mining this data alongside POI locations can provide significant practical benefits.

To bridge the scientific and real-world relevance of the algorithm, we introduce an enhanced P-KMEANS model. This is achieved by integrating POI seeds and refining the 'K' value and cluster center selection within the KMEANS framework. This model then conducts a cluster analysis of passenger behavior, resulting in life circles depicting travel patterns.

The P-KMEANS algorithm is designed to function based on a two-stage optimization principle. The first stage involves the analysis of the relationship between POIs and their corresponding travel implications, which aids in extracting a critical 'K' value associated with travel. This 'K' value is instrumental in defining central points for travel.

The second stage of optimization involves integrating the pre-determined 'K' value and central point into the K-Means algorithm, enhancing the algorithm's efficiency by strengthening the link between the clustering process and its inherent meaning. This fosters a more focused generation of clusters and boosts the overall performance of the algorithm.

The implementation of the algorithm model proceeds as follows:

(a) POI Data Acquisition: The POI spider capture algorithm(Stevanovic et al., 2012) (Utilizing web spider to extract POI categories from maps website) is utilized to gather current area POI data.

(b) Generation of POI Seeds: Mean-shift clustering computations are performed on the collected data to establish the initial 'K' value, which represents the number of clusters, and the initial central point of the clusters.

(c) Calculation of Clustering Results: The clustering results from Step 2 are used as input to conduct K-means clustering on people's travel trajectory data.

(d) Result Analysis: Multiple evaluation functions are employed to assess the outcomes of the clustering.

In the preceding segment, all POIs were established, and POI seeds were generated. Now, the count of POI clusters furnishes the 'K' input for the algorithm, with POI seeds serving as the initial centroid positions. Subsequently, we employ P-KMEANS clustering on bus travel trajectories. The pseudocode of the algorithm is delineated in Algorithm 1.

**Algorithm 1** P-Kmeans pseudocode

```
Input: K, Center, Data
Output: Clustering result
Initialize parameters
Create k points as the starting centroid (POI center point Center)
When the cluster allocation result of any point changes
  | For each data point in the dataset
  |   | For each centroid
  |   |   | Calculate distance between centroid and data point
  |   | Assign data points to their nearest cluster
for each cluster
  | calculate the mean of all points in the cluster and use the mean as the centroid
Return
```

The input parameters of the algorithm can be defined as follows: 'K' represents the count of clusters generated by the algorithm; 'Centre' signifies the initial clustering centre points; 'Data' denotes the collected set of travel trajectory data.

In summary, utilizing the P-KMEANS clustering method, we segregate a single passenger's travel trajectory into multiple clusters. These clusters coalesce to form a single passenger's travel life circle, providing a foundation for further analyses. This methodology effectively uncovers and highlights the passenger's distinct travel characteristics.

### *3.4.3. LDA Travel Feature Mining Based on POI*

In this section, we introduce the P-LDA algorithm specifically designed to extract and analyze passenger travel characteristics. In the preceding Section 3.4.2, we conducted a cluster analysis on passenger travel trajectories, dividing the passenger's travel objectives into several segments. The primary aim of this section is to calculate the characteristic weights of public transportation residents, utilizing both the POI data and passenger life circles. This is achieved by our proposed P-LDA model, which enhances the topic selection process of the LDA model.

Specifically, the P-LDA model enhances the LDA feature mining method by integrating POIs. Based on the passenger travel area segments identified in Section 3.3, we embark on a quantitative analysis to extract the feature weight ratio of passenger trips. Drawing upon the quantitative analysis from Section 3.3, the algorithm associates POI types with travel objectives. It treats the corresponding parameter values as observable variables and the passenger travel attributes as latent variables. Using the P-LDA model, we construct a joint probability distribution of passenger travel and resolve the unknown variables, thereby extracting passenger feature weights and gaining an intrinsic characteristic distribution of public transportation travel patterns. In Section 3.4.3.1, we conducted a travel clustering analysis and extracted travel characteristics. In Section 3.4.3.2, we further discussed and quantified these travel features.

#### *3.4.3.1. Clustering and Feature Mining*

Firstly, we conduct a cluster analysis of the various POIs within residents' life circles. The aim is to explore the relationships between different functional POIs in these life circles and their correlation with passengers' travel intentions. It is important to recognize that most public transport trips are driven by specific goals and are largely influenced by the functional attributes and attributes of urban structures. Passengers tend to favour routes that are in close proximity to their consumption destinations. The frequency of visits to a particular area, coupled with its distinct functional features, can serve as a reflection of individual passenger attributes. To illustrate this further:

- In mining passenger age attributes, teenagers primarily frequent educational POIs, middle-aged passengers are closely associated with work-related POIs, and elderly individuals are most connected to medical-related POIs.
- In mining gender characteristics, women are more likely to visit beauty salons, shopping centers, and service-oriented POIs, while men frequent entertainment and sports venues more often.
- When considering economic levels, Engel’s law can be applied. At lower income levels, food expenditures exceed other costs. Once basic needs are met, spending gradually shifts towards clothing, everyday necessities, travel, and so on.
- In mining occupational attributes, job nature can be inferred from workplace POIs. Office workers are predominantly located in corporate locations, party and government agency staff are mainly found in government agency POIs, and teachers are primarily in campus POIs.

To execute this analysis, we designate the cluster centre point of bus travel passengers as the epicentre of a circle, with a specific range serving as its radius. Subsequently, we enumerate all eligible POIs within this defined range. The distribution of POIs surrounding the passenger's trajectory is systematically recorded using a matrix, with classification labels sourced from Baidu map, as depicted in Table 3.

Table 3 POI industry classification labels.

| Category Name | Selected or not/Number | Record Name |
|---|---|---|
| Delicious Food | 0 | food ($I_{fo}$) |
| Hotel | 1 | hotel ($I_h$) |
| Shopping | 2 | shopping ($I_{sh}$) |
| Life Services | 3 | service ($I_s$) |
| Beauty | 4 | beauty ($I_b$) |
| Tourist Attractions | 5 | travel ($I_{tr}$) |
| Leisure and Entertainment | 6 | entertainment ($I_{en}$) |
| Fitness | 7 | sports ($I_{sp}$) |
| Educational Training | 8 | education ($I_e$) |
| Culture and Media | 9 | media ($I_{me}$) |
| Medical Treatment | 10 | medicine ($I_m$) |
| Car Services | 11 | car ($I_{ca}$) |
| Traffic Facilities | 12 | traffic ($I_t$) |
| Financial | 13 | finance ($I_f$) |
| Property | 14 | estate () |
| Incorporated Business | 15 | company ($I_c$) |
| Government Institutional | 16 | government ($I_g$) |
| Entrances and Exits | Nonuse | - |
| Natural Features | Nonuse | - |
| Administrative Landmark | Nonuse | - |
| Gate Address | Nonuse | - |

To augment the accuracy of our investigation, this study selectively focuses on POI categories that bear significant correlation with daily human activities, systematically excluding those that demonstrate a low level of correlation such as entrances and exits, natural features, administrative landmarks, and gate addresses.

Using the centre point of a passengers' trajectory as the centre of a circle and the travel distance (DIS) as the radius, we count all POI points within this circle and categorize them based on their classification. The labelling formula is as Eq. (5) :

$$M_{k,t} = \sum_{i=0}^{N_t} \left( M_{k,t,i} + \left( 1 - \frac{dis_i}{DIS} \right) \right) \tag{5}$$

where $M_{k,t}$ denotes the POI distribution of class $t$ in the k-th cluster centre point circle; $N_t$ represents the number of POIs of class $t$; $M_{k,t,i}$ denotes the i-th data point in the k-th cluster centre class of $t$; $t$ signifies the classification label to which the POI point belongs; $dis_i$ represents the distance between the POI points and the cluster centre; and DIS is the search radius of the POI.

A POI point contributes more to its category when it is nearer to the centre point. We traverse the POI points surrounding the centre point being processed and assign weights to the vector items, thereby expressing the proportion of various POIs around the passenger point as a percentage. The calculation formula is as Eq. (6):

$$M_{k,t} = \frac{M_{k,t}}{\sum_{i \in TAG} M_{k,i}} \tag{6}$$

For a single passenger, the sum total of each POI class in his life circle is calculated, combined with the number of tracks $N_k$ in the life circle. Thus, the POI distribution in the life circle of the passenger is obtained

and defined as Eq. (7):

$$X_{i,t} = \sum M_{k,t} * \sqrt{N_k} \tag{7}$$

where $X_{i,t}$ denotes the POI distribution of the i-th passenger in category $t$; where $N_k$ signifies the number of trajectory points contained within the cluster corresponding to the k-th cluster centre point.

Given that the number of clusters generated by passenger trajectory points is not uniform and that individual datasets may be large, thus significantly influencing experimental results, we perform arithmetic square root calculations on $N_k$ to mitigate the effects of quantitative differences arising from cluster partitioning. Furthermore, to maintain consistency in the passenger model's weight during attribute mining, the data vector $X_i$ is proportionally scaled so that the sum of each item is fixed. Consequently, the bus travel mode matrix $X$ based on POI is generated.

*3.4.3.2. LDA Feature Mining Based on POI*

This section, utilizing POIs and life circles (calculated based on 3.4.3.1), optimizes the topic selection process of the LDA model. It aims to extract the proportion of various passenger travel characteristics. The LDA is a Bayesian-based unsupervised clustering algorithm extensively utilized for clustering, identification, analysis, and mining of latent topic information in document collections or corpora. LDA leverages the bag-of-words(Klopries and Schwung, 2023) approach for information extraction, focusing only on the occurrence and frequency of words in a document, while ignoring the sequence of their appearance.

This generative mechanism of LDA model is illustrated in Fig. 4. a. In the LDA model, where $K$ represents the number of topics within a single document, where $M$ stands for the number of documents in a corpus, and $V$ indicates the vocabulary size. Where $M_k$ signifies the weighted representation of $K$ topics within a document; Where $V_k$ stands for the weighted representation of $V$ terms within a topic. The hyperparameters for and are $\alpha$ and $\beta$, respectively, according to the Dirichlet distribution. LDA, when analysing text, processes the source document using the bag-of-words approach, disregarding the sequence of words and concentrating solely on their frequency of occurrence.

In contrast to the conventional LDA model, the P-LDA model conceptualizes each POI classification label as a word in a document, a single bus travel pattern as a document in a corpus, and all matrices of bus travel patterns as a corpus. Hence, the bus travel mode can be perceived as a blend of different transport patterns, with the POI classification label reflecting the mapping of these patterns. To explore passenger attributes across diverse categories, this study opts for POI classification labels most correlated with distinct categories as the vocabulary. When mining age attributes, the POI tags for education, employment, and medical can be selected due to their significant association with adolescent, middle-aged, and elderly demographic groups, while other less influential POI tags can be excluded.

The enhanced generative mechanism of the P-LDA model is illustrated in Fig. 4 (b). In this model, $K$ signifies the number of passenger characteristic attribute categories; $M$ represents the number of passengers; $V$ stands for the POI classification label; $\theta$ indicates the proportion of residents' bus travel attributes; $\phi$ demonstrates the mixed weight of the various POI class tags in each attribute. Both $\theta$ and $\phi$ follow a Dirichlet distribution, with hyperparameters $\alpha$ and $\beta$.

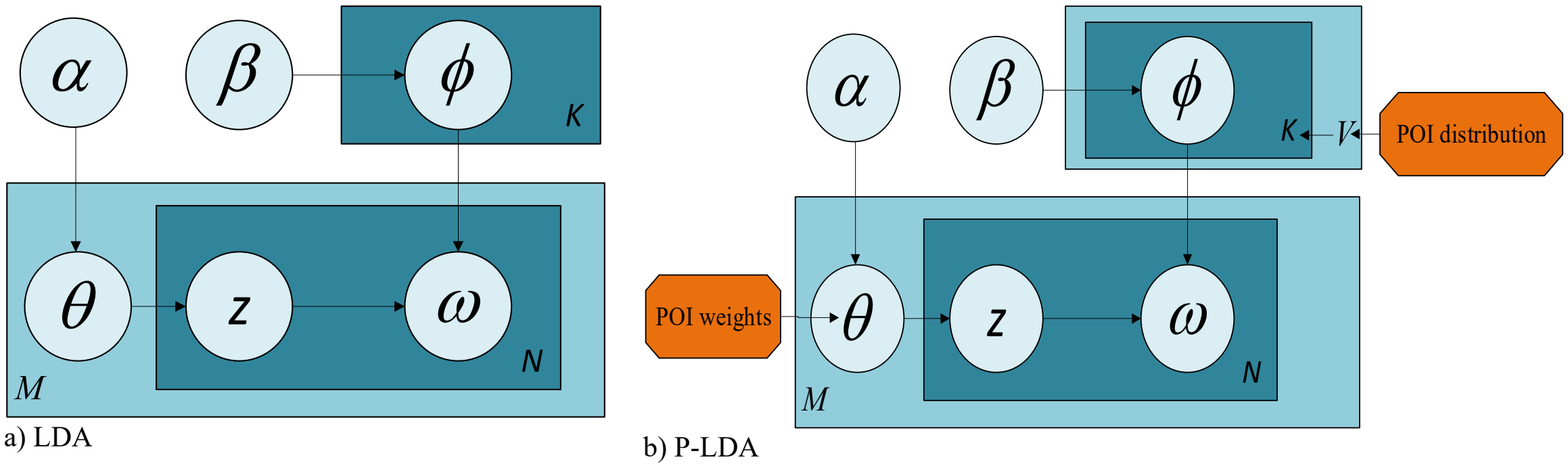

Fig. 4 Generation mechanism of LDA and P-LDA algorithms.

The feature generation process unfolds as follows:

(a) The topic library contains topics $\{1, \dots, K\}$. For any given $k \in \{1, \dots, K\}$, a distribution vector $\phi_k \sim$ Dirichlet( $\beta$ ) distribution.

(b) The data sample comprises passengers $\{1, \dots, M\}$. For any $m \in \{1, \dots, M\}$, there's a corresponding distribution vector:$\theta_m \sim$ Dirichlet($\alpha$).

(c) The POI classification encapsulates labels $\{1, \dots, N\}$. For any $n \in \{1, \dots, N\}$, the topic is represented as Zn ~ Multinomial ($\theta_m$) distribution, where $Z_n \in \{1, \dots, K\}$.

(d) Upon selecting a feature attribute $Z_n$, it can be symbolized as $Wn \sim$ Multinomial($\phi_{zn}$) distribution, where $Wn \in \{1, \dots, V\}$.

In summary, the P-LDA algorithm is employed for topic classification based on the results of cluster mining as per Section 3.4. Through the feature analysis, the distinctive attributes of various passengers' trips and their weight ratios are methodically extracted.

## 4. Case Study

Building on the methodology from Section 3, this section will apply these enhancement techniques to specific real-world scenarios, deliberating on the practical applicability of the method and its potential contribution to bus travel feature extraction. Section 4.1 describes the data source and structure used in the case study. In Section 4.2, we implement the improved mining method to delineate bus travel attributes. Section 4.3 evaluates the performance of this enhanced method, and finally, Section 4.4 offers a summary and analysis of the extracted bus travel characteristics, providing a holistic understanding of the results.

### *4.1. Data Source and Data Processing*

In this study, we primarily leverage two types of data: POI data derived from maps and bus travel trajectory data. The map API was used to capture the former, while the latter was extracted from the Yi-bus database(Beijing Zhonghang Xun Technology Co., 2017-01-22). Subsequent sections provide a more detailed breakdown of the data acquisition and processing strategies implemented.

#### *4.1.1. POI data*

For the foundation of this study, we utilized Baidu Map's (Baidu Net Science and Technology Co., 2016-02-14) API to mine POI data within Yan'an's urban region in China. This extracted data forms an integral part of the input for subsequent stages of our investigation. In Section 4.1.1.1, we collected POI data using web crawlers. In Section 4.1.1.2, we conducted data preprocessing.

#### *4.1.1.1. Data capture*

Employing the Baidu Map API, we systematically categorized and gathered data across Yan'an City's vast expanse of 3,556 square kilometres. The volume of data points fluctuates depending on the types of collections. For categories like ["food," "foreign restaurant," "snack fast food restaurant," "cake dessert shop," "cafe," "tea shop," and "bar"], our data collection effort encompassed 20 pages. This compilation of data was completed on December 20, 2021.

Throughout this data collection phase, we accumulated a total of 349 samples. Examples from this collected sample data are presented in Table 4:

Table 4 Sample data from the Baidu Map.

| | Latitude | Longitude | Name | Label | City | Area | Address |
|---|---|---|---|---|---|---|---|
| 1 | 36.628798 | 109.436867 | Old Village Chief Iron Pot Stewed Lamb (Zaoyuan Park) | Delicious Food | Yan’an | baota | In the parking lot of Zaoyuan Park, Zaoyuan Road, Baota District, Yan'an City, Shaanxi Province |
| 2 | 36.603322 | 109.492363 | Four Seasons Salt and Taste (Center Street Store) | Delicious Food | Yan’an | baota | 4th Floor, Huiyuan Hotel, No. 18, Fenghuang Street, Baota District, Yan'an City, Shaanxi Province |
| 3 | 36.632371 | 109.434262 | Yan'an Eight Big Bowls (First Branch) | Delicious Food | Yan’an | baota | Next door to Shengli Primary School, Guanghua Road, Nanshi Street, Baota District, Yan'an City, Shaanxi Province |
| … | … | … | | … | … | … | … |

#### *4.1.1.2. Data processing*

Due to the potential variances in data classification methods, we might encounter instances of repeated data points. For instance, a "snack fast food restaurant" could be classified under both the "food" and "foreign restaurants" categories, thereby leading to data duplication. Hence, before progressing to the experimental stages, it's crucial to manually curate the data to label null entries, remove repeated data, and correct any misclassified data.

The collected data is then converted into a Data Frame format leveraging the panda’s library (Gavriilidis et al., 2023). Subsequently, it's transformed into a list format, enabling us to expunge duplicate entries, null data, and corrupted data. Nonessential columns such as city and region are discarded, and the data is vectorized. After implementing these processing techniques, we're left with a data set comprising 286 items. The formatted data is exhibited in Table 4.

### *4.1.2. Travel trajectory data*

The trajectory data used in this section is sourced from the trajectory mining data results discussed in Section 2. This data set is constituted of bus travel data from Baota District in Yan'an City, gathered from January to June 2018 using specialized software. Initially encompassing 922,387 data points, the data set, after processing, consists of 850,720 data points. The representation of this data is demonstrated in Table 5:

Table 5 Trajectory data structure.

| | uid | lng ($T_l$) | lat ($T_d$) | up_time ($T_t$) |
|---|---|---|---|---|
| 1 | 1EE2…0F7A68 | 109.515776 | 36.613244 | 2018/2/1 11:23:56 |
| 2 | 0819…E45B7B | 109.413065 | 36.628334 | 2018/2/10 8:01:12 |
| 3 | 1550…9BF2F3 | 109.489039 | 36.595997 | 2018/2/23 20:17:04 |
| 4 | 4540…8B981B | 109.515826 | 36.612331 | 2018/2/27 18:43:33 |
| 5 | … | … | … | … |

The current data points are disorganized and lack connections, necessitating structuring via the unique identification code, Uid. By dividing the data using Uid, we can collate the travel data points of each individual into a separate file. Our research indicates that the number of trajectory records varies among different passengers. Approximately 66.7% of the total dataset consists of passengers with fewer than 100 trajectory records, whereas passengers with more than 100 records constitute about 0.5% of the total. The statistical results are depicted in Fig. 5:

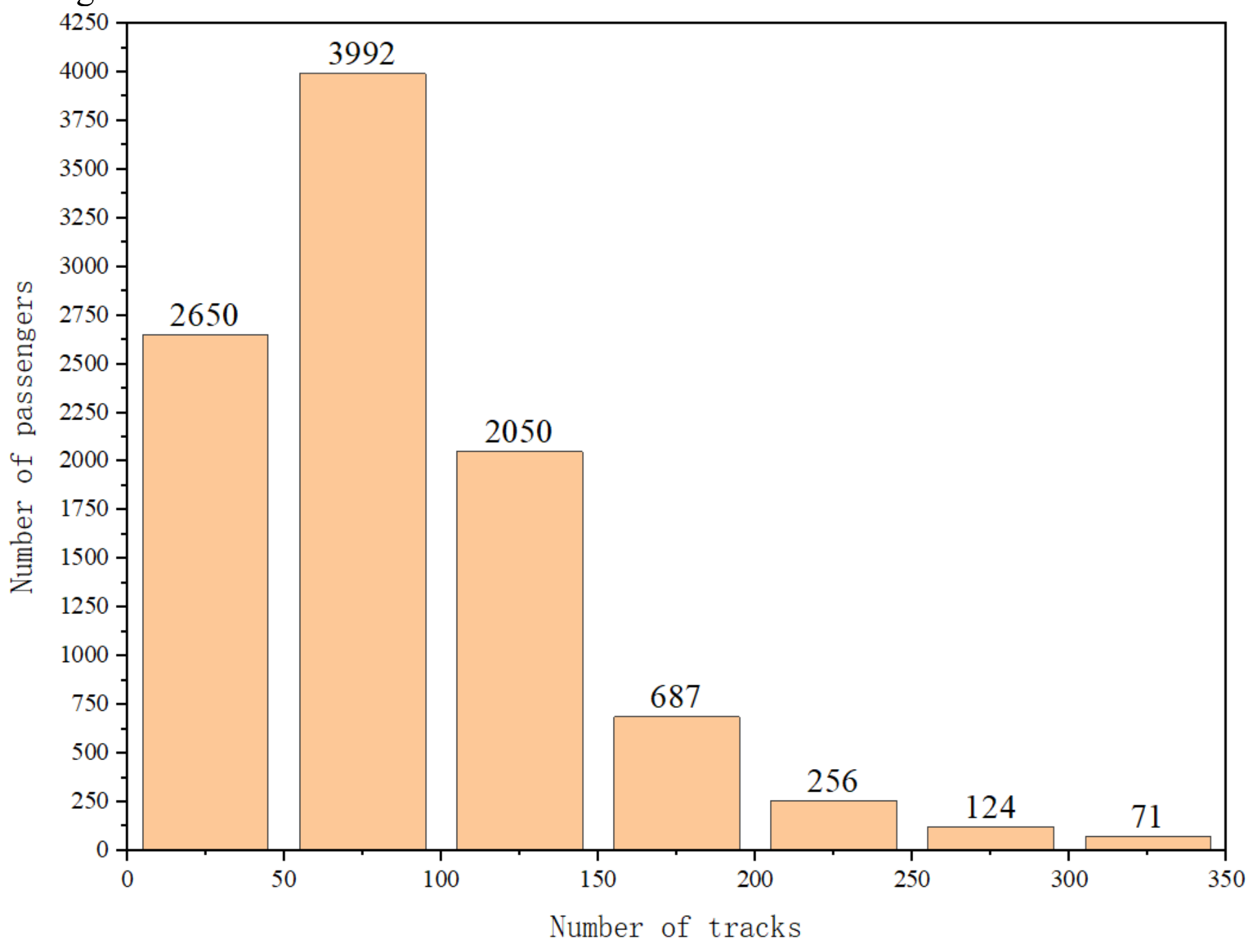

Fig. 5 Statistical results of the number of bus trip records.

Given that the conclusions on potential attributes of passengers in the study are derived from the travel trajectory records, we posit that passenger trajectory data with a larger number of records will more accurately reflect passenger characteristics. Therefore, the study filters out passenger data with more than 100 records as the passenger trajectory dataset for research use. This approach will augment the reliability of the research findings and enhance computational efficiency. In later stages, further comparisons will be conducted to select the most suitable passenger dataset for the research data source.

*4.2. Mining and analysis of bus travel characteristics*

Based on the travel trajectory data discussed in Section 4.1, this section integrates POI seeds into cluster mining and feature discovery in order to analyze and excavate bus travel characteristics. In Section 4.2.1, the KMEANS clustering algorithm is improved by pre-generating POI seeds. This enhancement aims to construct a P-Kmeans clustering model that not only improves the efficiency of trajectory mining but also produces clustering results that better align with human travel behavior. In Section 4.2.2, the LDA feature mining algorithm is further improved by integrating POI seeds. The goal is to build a P-LDA feature mining model that, utilizing the resident travel life circle generated by the clustering in Section 4.2.1, summarizes and mines the intrinsic weight attributes of passenger travel. This process is conducted from a quantitative analysis perspective, allowing for a more comprehensive understanding of passenger travel behavior.

*4.2.1. P-KMEANS feature clustering*

In Section 4.2.1.1, we generated POI seeds. In Section 4.2.1.2, we produced K-MEANS clustering results.

*4.2.1.1. POI seed generation*

The Mean-Shift algorithm creates clusters by iteratively moving the centre of the clustering circle towards the area with the highest cluster density until an optimal cluster is formulated. The most densely populated point within the dataset becomes the centre of each respective cluster. In this study, the Mean-Shift algorithm is harnessed to cluster POIs, leading to the formation of five distinct clusters. The centres of these clusters are denoted as marks. The results are graphically represented in Fig. 6.

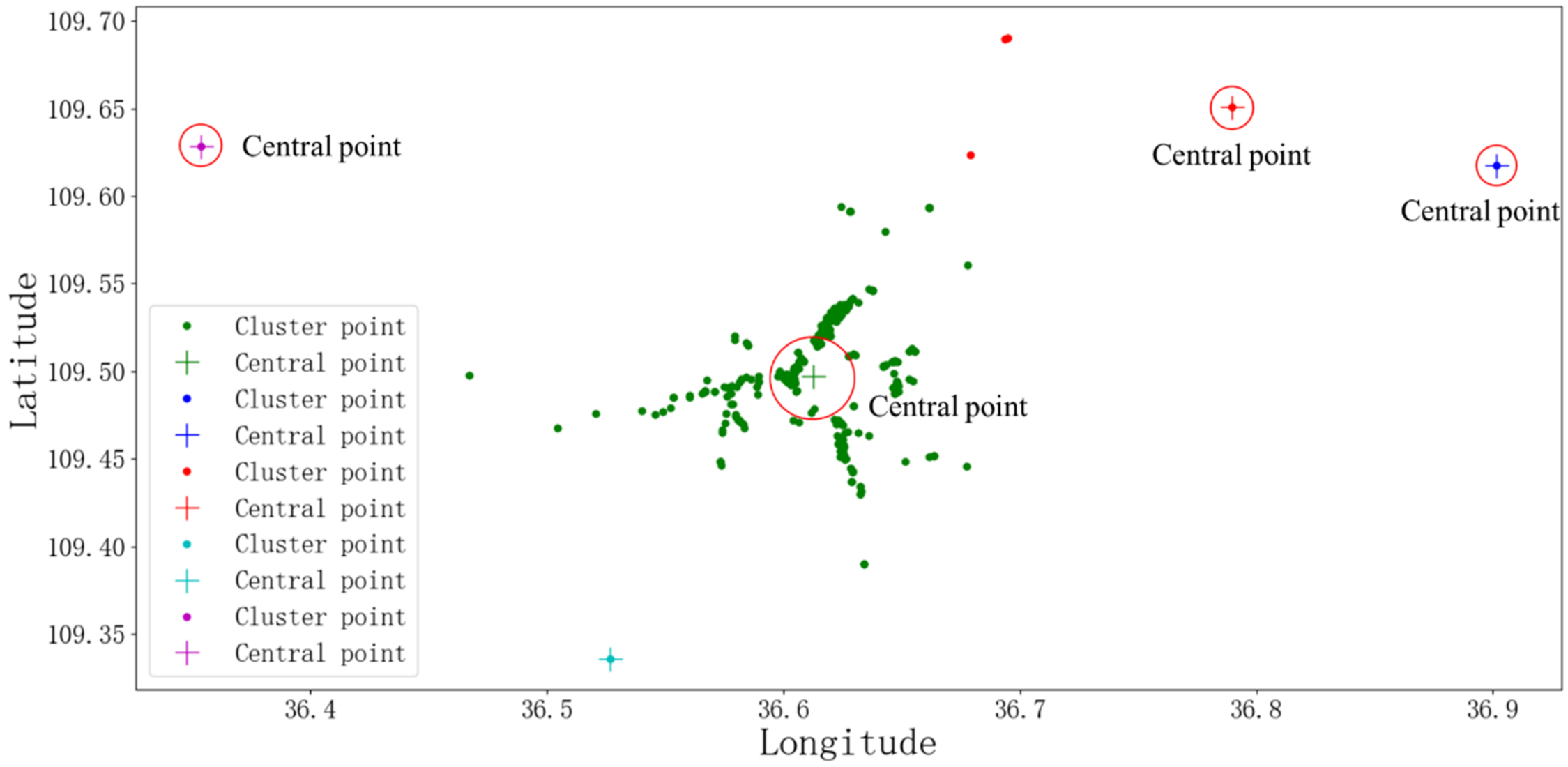

Fig. 6 MeanShift seed generation results.

As evident from the results, each category of POI exhibits a unique distribution on the map. POIs of the same type naturally coalesce into clusters, each representing a distinct functional area within the region. Moreover, the center of each cluster signifies the concentration of similar POIs within a given region, effectively functioning as the nucleus of the POI functional block.

### *4.2.1.2. P-KMEANS cluster generation*

Employing the previously established POI seed as the initial centre point for KMEANS clustering, and taking into account the number of clusters produced, passengers with 640 trajectory points are selected as the initial dataset for this experiment. The results of the clustering are visually presented in Fig. 7.

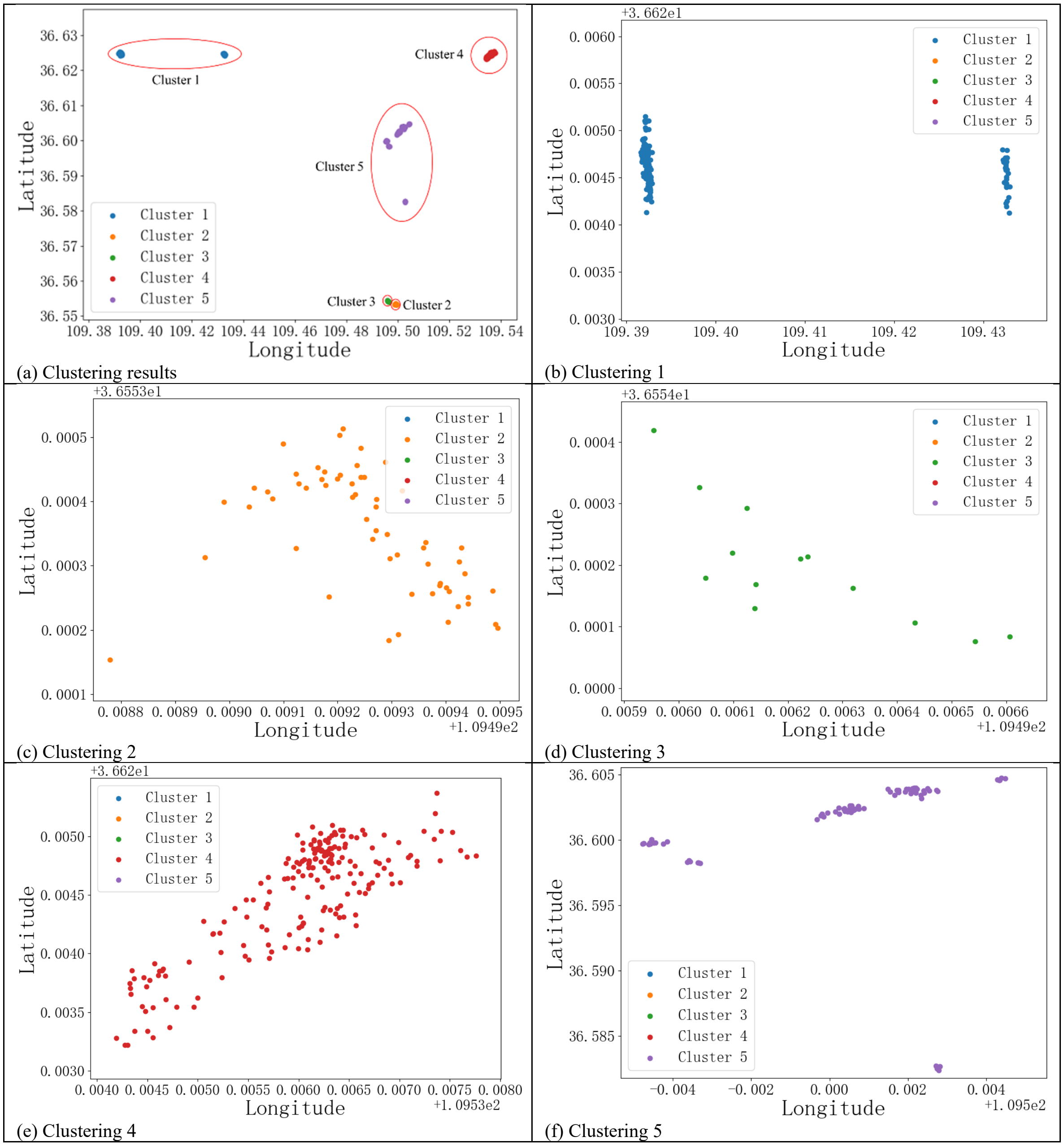

Fig. 7 Food P-Kmeans clustering results.

The P-Kmeans clustering, leveraging the existing clustering seed, yields a total of five clusters. These clusters respectively represent activity trajectories centred around food in the passengers' current activity patterns. By carrying out a multi-POI seed clustering analysis on passengers, we can infer the passengers' preferences and frequency of visits to different POIs. This process also aids in identifying the cluster centres of various clustering results.

The outcomes derived from seed-based clustering elucidate the 'life circle' fashioned by passengers around the seed given their current mode of travel. This encompasses the 'gourmet life circle' that materializes when 'food' is the seed, and the 'travel life circle' when 'tourism' acts as the seed. The number of activity trajectories indicates the frequency of passengers appearing in the current seed environment and their resemblance to the POI centres.

*4.2.2. P-LDA feature mining*

Building upon the previous section, we now focus on examining and interpreting the nuances of urban living circles, which are commonly defined by walking distances. As noted by Su T et al. (Su, 2021), resident living circles are segmented based on the travel distance and time for different passenger groups and the service radius of various public facilities. In this context, the smallest extent of a basic living circle is typically defined as a 15 to 30-minute walk, which equates to roughly 500 meters. Consequently, in this study, the cluster centre of the bus travel trajectory is utilized as the circle's centre, and the 500-meter range is employed as the circle's radius to map out all POIs within the scope.

Through P-KMEANS, we generate several core trajectory centre points based on the bus travel trajectory. We then use the trajectory centre point as the centre, and count the radius within 500 meters - this forms the coverage area of residents' activities in the itinerary. Subsequently, we used P-LAD model to tally the POI points in this activity domain and perform weight calculations for data analysis. The resulting calculations are presented in Table 6.

Table 6 POI weight calculation results

| UserID | food | hotel | shop | serv | beau | trav | ente | spor | educ | media | … |
|---|---|---|---|---|---|---|---|---|---|---|---|
| 1 | 5 | 3 | 7 | 17 | 12 | 0 | 3 | 3 | 0 | 1 | … |
| 2 | 2 | 0 | 18 | 9 | 22 | 1 | 1 | 0 | 27 | 3 | … |
| 3 | 3 | 5 | 23 | 17 | 13 | 4 | 6 | 23 | 16 | 32 | … |
| 4 | 8 | 21 | 34 | 28 | 12 | 8 | 9 | 11 | 57 | 11 | … |
| 5 | 10 | 3 | 12 | 16 | 0 | 5 | 1 | 0 | 0 | 21 | … |
| … | … | … | … | … | … | … | … | … | … | … | … |

In this table, "food" represents the POI point for food-related activities, "hotel" symbolizes the characteristic value of accommodation POIs, and "shop" stands for the characteristic value of shopping POIs. Likewise, "serv", "beau", "trav", "ente", "spor", "educ", and "media" represent the characteristic values of life service, beauty, tourist attraction, leisure and entertainment, sports and fitness, education and training, and cultural media POIs, respectively.

The figures in the table correspond to the computed characteristic values of each type of POI. Different passengers possess distinct characteristic weight proportions, leading to diverse travel probabilities. For instance, 'User 1' has 'Service' as its most heavily weighted characteristic, implying that this passenger's travel purpose is primarily service-oriented, followed by other attributes. By leveraging these feature weights, the distinct personality traits of various passengers can be effectively discerned.

## *4.3. Performance Evaluation*

Building on the outcomes of cluster mining and feature discovery detailed in Section 4.2, this section employs various evaluation metrics to demonstrate the enhanced performance of the KMEANS and LDA algorithms when POI seeds are incorporated, as illustrated through experimental data. In Section 4.3.1, we conducted a performance evaluation of the P-KMEANS method. In Section 4.3.2, we evaluated the performance of the P-LDA method.

### *4.3.1. P-Kmeans Performance Evaluation*

The generated clusters are evaluated using three indices: the silhouette score, the Calinski Harabasz score, and the Davies-Bouldin score. 1) The Silhouette Coefficient Index combines cohesion and separation to assess the impact of various algorithms or their operational modes on clustering out-comes from identical data. 2) The Calinski Harabasz Index, is the ratio of inter-cluster dispersion to intra-cluster dispersion. 3) The Davies-Bouldin Index, proposed by David L. Davies and Donald Bouldin, assesses the quality of clustering algorithms.

The resulting evaluations are depicted in Fig. 8.

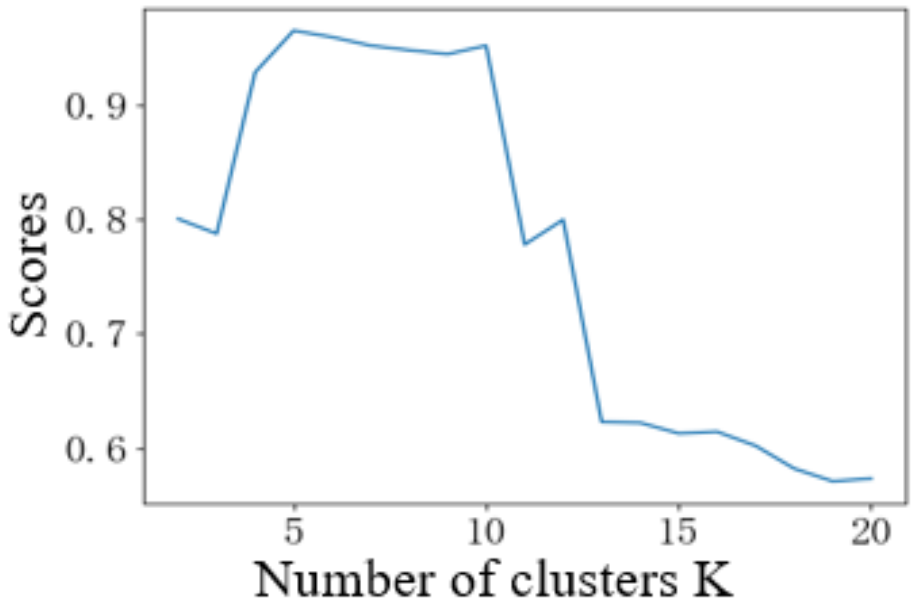

(a) Silhouette Coefficient

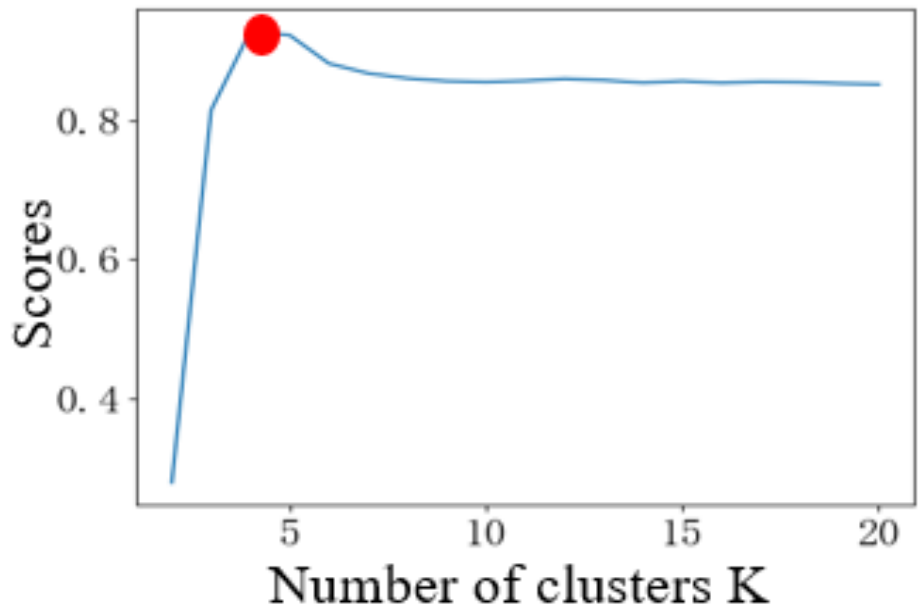

(b) Silhouette Coefficient

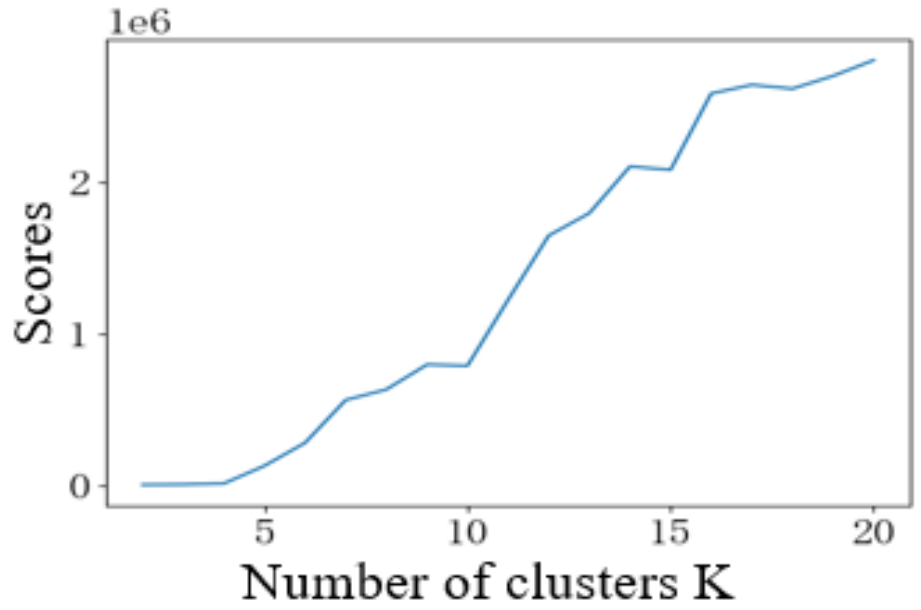

(c) Calinski Harabasz

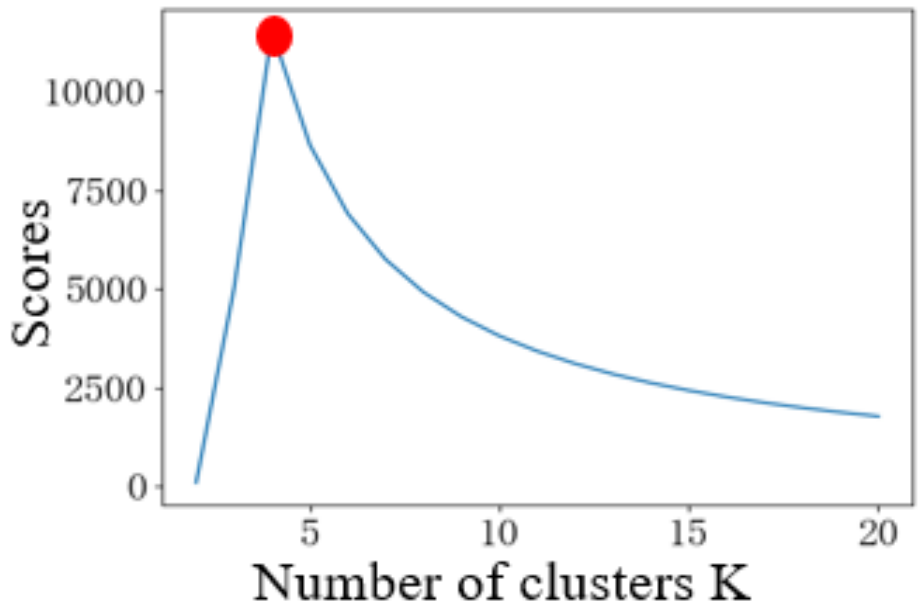

(d) Calinski Harabasz

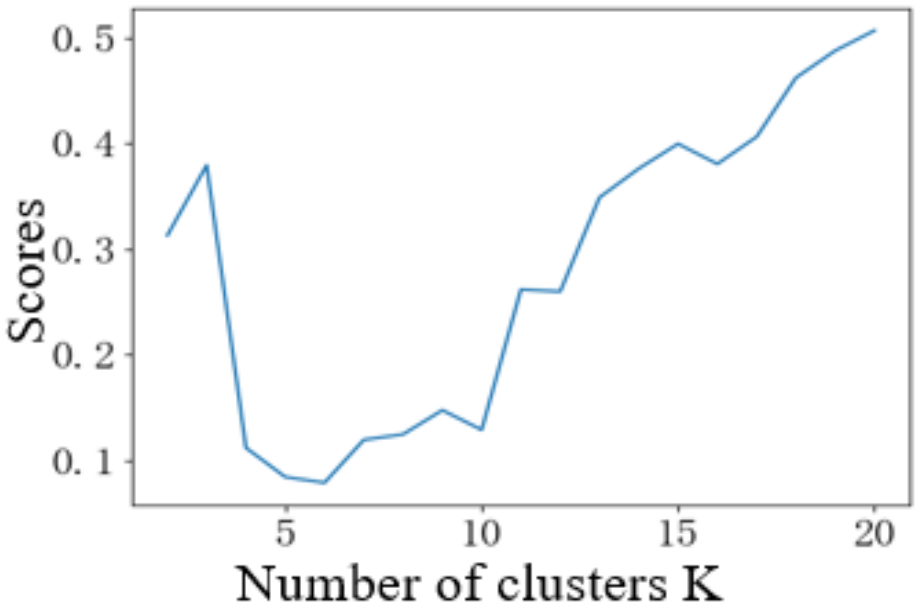

(e) Davies Bouldin

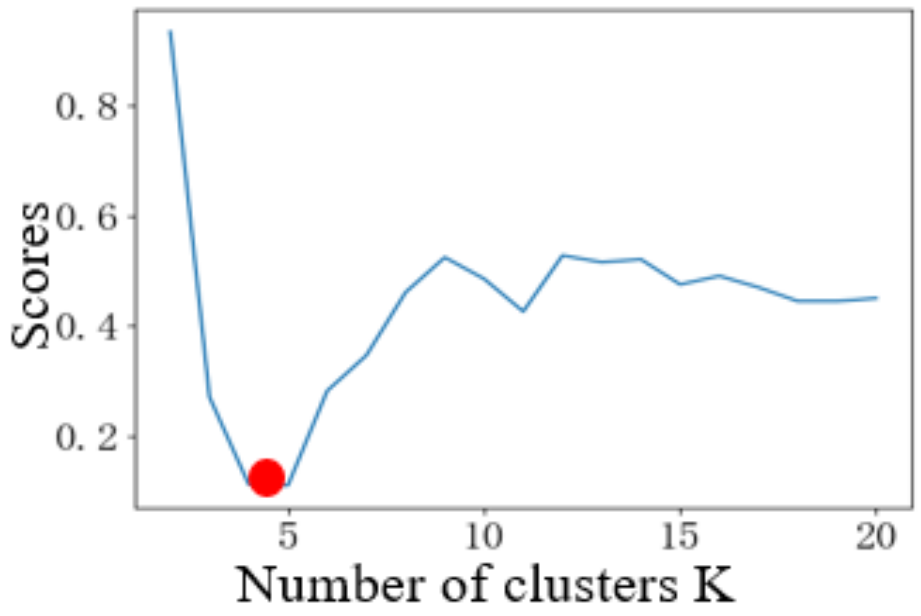

(f) Davies Bouldin

Fig. 8 P-Kmeans performance evaluation results. (a) Silhouette score index of clustering generated for unspecified initial POI seeds; (b) Silhouette score index of clusters generated for specified initial POI seeds; (c) Calinski-Harabaz index for clustering generated by unspecified initial POI seeds; (d) Calinski-Harabaz index for clusters generated for specified initial POI seeds; (e) Davies Bouldin Index for clusters generated for unspecified initial POIs; (f) Davies Bouldin Index for clusters generated with specified initial POI seeds.

The evaluation of clustering by the Silhouette Score Index indicates that the optimal number of clusters (K value) is 5, irrespective of whether the initial POI seed is specified. However, when the POI seed is defined, the resultant clustering curve presents greater uniformity, thereby improving the execution efficiency and providing a more stable and reliable outcome.

An examination through the lens of the Calinski Harabasz Index's clustering evaluation suggests that without a defined initial POI seed, the clustering process fails to effectively converge, which hinders the attainment of an optimal K value. Conversely, when the POI seed is specified, the process converges efficiently to a K value, yielding a result that concurs with the Silhouette Score Index.

However, the Davies-Bouldin Index's evaluation introduces a discrepancy. In the absence of a predetermined initial POI seed, the optimal clustering is achieved at K=6, a result that stands in contrast with those from the Silhouette Score Index and Calinski Harabasz Index. Nonetheless, when the POI seed is specified at initialization, the optimal clustering value of K=5 is found, bringing it into alignment with the other two evaluations.

In conclusion, the experimental results highlight that predefining POI seeds can significantly augment the algorithm's execution efficiency, minimize its time complexity, and reliably produce optimal clustering results.

### *4.3.2. P-LDA Performance Evaluation*

The efficacy of the algorithm is scrutinized through experiments using the Yan'an Yi-bus dataset, which comprises 9961 passengers and 850,720 trajectories. To enhance the efficiency of sample analysis, we restricted our initial examination to passengers with over 100 trajectories, which yielded a total of 7311 passengers. We partitioned the obtained passenger trajectories randomly in a ratio of 80-20. The majority, 80%, served as training data to instruct the P-LDA model, while the remaining 20% were used as validation data. The performance of the P-LDA model is demonstrated via Recall, Precision, and F1-Score, which together illustrate its adeptness in mining feature attributes. The definitions of these indices are available in (Goutte and Gaussier, 2005).

Through the LDA mining analysis outlined in Section 3.3, it was discerned that each passenger's P-LDA model encompasses seven topics: age, occupation, gender, health, economic, safety, and personality. The weights assigned to each topic within a passenger's model are not uniform, leading to variance in the model's emphasis on different topics. For this reason, to gain predictive insight, a series of controlled experiments were carried out across a diverse set of passengers.

An array of methods including K-means(Chabchoub and Fricker, 2014), DBSCAN(Density-Based Spatial Clustering of Applications with Noise)(Yao et al., 2022), spectral clustering(Sharma and Seal, 2021), ST-RNN(Spatial-Temporal Recurrent Neural Network)(Wang et al., 2023), HCA(Hierarchical Cluster Analysis)(Zhang et al., 2018), LLE(Locally Linear Embedding)(Nichols et al., 2011), and T-SNE(t-distributed stochastic neighbor embedding)(Kuntalp and Düzyel, 2024) were employed to calculate the Recall, Precision, and F1-Score for the passengers' original trajectory data. The findings of these calculations are tabulated in Table 7.

Table 7 Multi-algorithm evaluation and comparison results.

| Algorithm | Recall | Precision | F1-Score | MAE |
|---|---|---|---|---|
| DBSCAN | 0.086 | 0.021 | 0.079 | 0.278 |
| Spectral Clustering | 0.117 | 0.088 | 0.114 | 0.154 |
| ST-RNN | 0.164 | 0.117 | 0.101 | 0.137 |
| HCA | 0.064 | 0.058 | 0.008 | 0.095 |
| LLE | 0.092 | 0.039 | 0.095 | 0.056 |
| T-SNE | 0.126 | 0.106 | 0.033 | 0.041 |
| KMEANS | 0.152 | 0.133 | 0.223 | 0.247 |
| P-LDA | 0.213 | 0.169 | 0.147 | 0.321 |

The experimental results reveal that the P-LDA algorithm exhibits superior performance compared to other algorithms with respect to Recall, Precision, and Mean Absolute Error (MAE). The K-means algorithm does marginally exceed the P-LDA algorithm in terms of F1-Score. Nevertheless, it is important to note that a P-K-means operation is conducted on the data during the calculation of the POI seed, which marginally inflates the F1-Score value. Despite this, the P-LDA model consistently shows a higher level of predictive performance across the majority of key metrics. Fig. 9 provides a clear illustration of the experimental results, demonstrating the superiority of the P-LDA algorithm in terms of predicting most attributes over other algorithms.

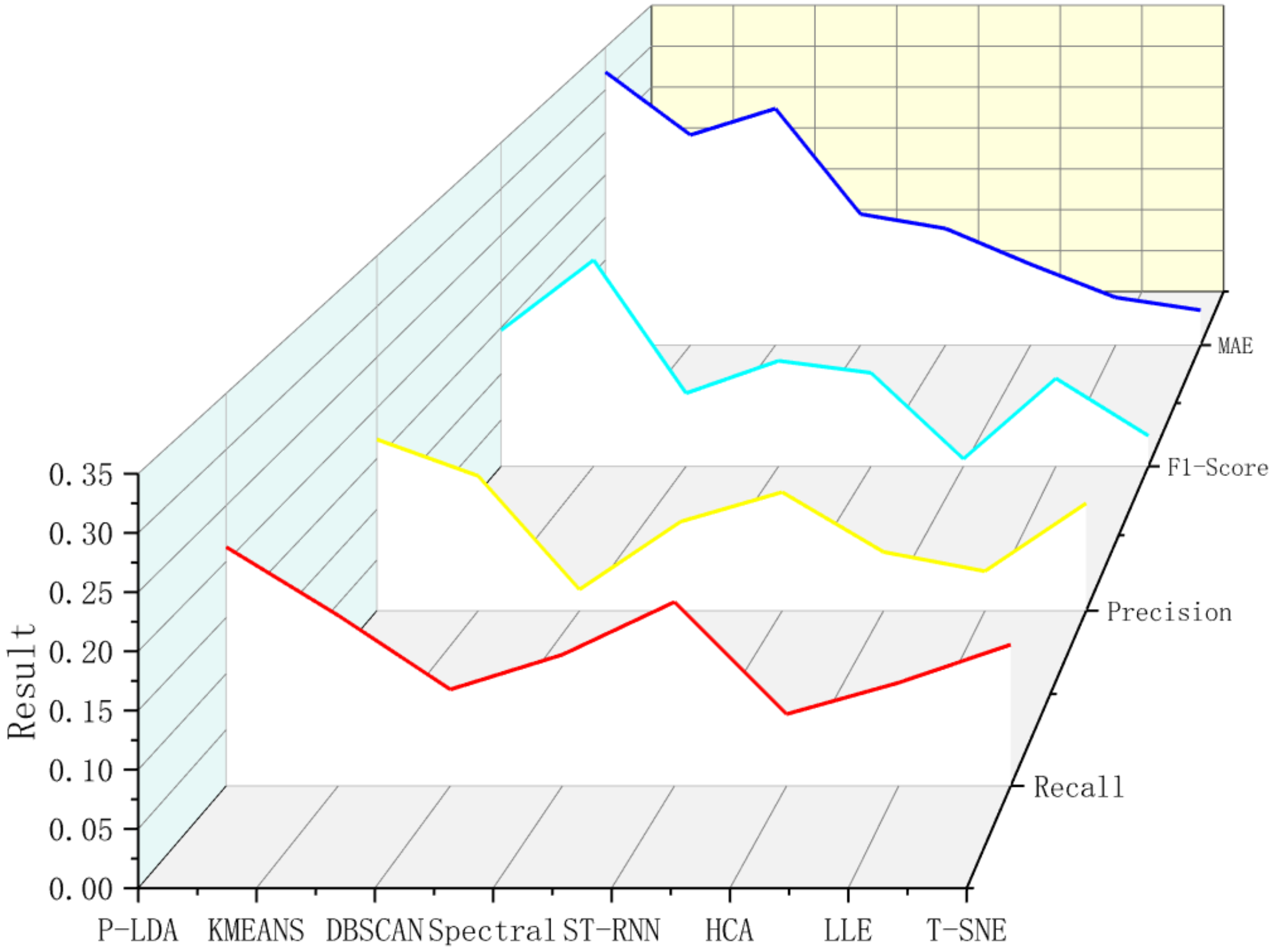

Fig. 9 Experimental comparison results.

### *4.4. Analysis of bus travel characteristics*

The bus mode of transportation represents a blend of multiple latent feature attributes, with the POI weight offering a tangible reflection of these attributes. Mining the travel POI feature weight allows for an analysis of a passenger's potential feature attributes and facilitates predictions regarding future travel patterns using this data. A passenger's mode of travel is influenced and characterized by a variety of unique attributes. Consequently, when conducting feature analysis, it is crucial to isolate primary POI feature attributes, discard classifications with negligible impact, and use the selected feature pattern matrix as the corpus in the LDA model.

To further explore public transport optimization, this study focuses on the attributes most closely associated with public transportation: age, occupation, gender, sports, cost, safety, and personality characteristics. By analysing the attributes of a POI, each attribute of a feature can be extracted. The passenger feature attributes correspond to the classification labels of POIs, and their representation method is correlated with the classification weight. During computations, the impact of secondary factors is ignored, while only the influence of primary factors is considered. The criteria for judgement are detailed in Table 8.

Table 8 Judgment of POI tags and classification corresponding to passenger feature attributes.

| Feature attribute | POI classification label selection | Number of classifications and classification results | Result judgment |
|---|---|---|---|
| Age ($P_a$) | company, government, education, medicine, traffic, car | 3(1 teenagers, 2 middle age, 3 old age) | Teenagers: education, traffic |
| | | | Middle age: company, government, car |
| | | | Old age: medicine, traffic |
| Occupation ($P_o$) | education, media, medicine， service, company, government, finance | 5 (1 students/teachers, 2 company employees, 3 political parties, 4 self-employed/businessmen, 5 retirees/medical workers) | Student/Teacher: education, media |
| | | | Retired/medical workers: medicine |
| | | | Company staff: service, company |
| | | | Party organ: government |
| | | | Self-employed/businessmen: finance |
| Gender ($P_g$) | shopping, beauty, service, traffic, car, entertainment, sports | 2 (1 male and 2 female) | Male: car, entertainment, sports |
| | | | Female: shopping, beauty, service, traffic |
| Health ($P_h$) | sports， entertainment， travel | 4(1 strong like, 2 like, 3 neutral, 4 dislike) | Better: sports |
| | | | Worse: entertainment, travel |
| Economic ($P_e$) | food, shopping, travel, entertainment, media, traffic, car | 5(1 cost insensitive, 2 moderately sensitive, 3 moderately sensitive, 4 sensitive, 5 very sensitive) | Better: shopping, travel, entertainment, media, car |
| | | | Worse: traffic, food |
| Safety ($P_s$) | sports, medicine, service, traffic, entertainment | 4 (1 lower, 2 intermediates, 3 higher) sense of security | Better: sports, medicine, traffic, service |
| | | | Worse: entertainment |
| Personality ($P_p$) | shopping, entertainment， media, travel, sports, service | 3(1 quiet, 2 average, 3 like noisy) | Better: service, sports |
| | | | Worse: shopping, media, entertainment, travel |

In terms of age characteristics, different age groups are associated with different types of POIs. For instance, teenagers frequently visit educational institutions, while young adults such as office workers primarily visit corporate and government buildings. Elderly individuals often frequent healthcare and wellness facilities. Middle-aged individuals show a preference for self-driving, thus frequenting car-related POIs more than other age groups. In contrast, teenagers and elderly individuals are more likely to rely on public transport, leading to a higher frequency of visits to transport facility POIs.

Occupational attributes significantly influence POI affiliations. For example, enterprise POIs are frequently visited by individuals employed within them. Students and educators often associate with training and cultural media POIs, while individuals in the labour force may frequent exhibition halls and art galleries. Workers in life services often visit business halls, post offices, logistics companies, and intermediaries. Medical staff are commonly associated with hospitals and medical institutions, and government employees are likely to visit government-related POIs. For retirees, POI visits can be predicted based on their age category and frequency of visits to healthcare POIs.

Gender attributes also play a role in shaping POI visitation patterns. Women tend to frequent beauty salons, shopping centres, and life service locations, while men often visit sports and entertainment-related POIs. As for transportation preferences, women are more likely to choose public transportation or walking, whereas men are more likely to drive or rent vehicles. Thus, women are observed to visit transport facility POIs more often, and men are more likely to be found at car service centres and parking lots.

Sports features can be inferred from the frequency of visits to sports and fitness POIs. The amount of time dedicated to sports activities typically falls outside of work or school hours and may be impacted by other leisure activities, such as visits to entertainment and tourism POIs.

Regarding cost characteristics, an individual's economic situation influences their lifestyle costs, which can be inferred from daily consumption patterns. For instance, those with a lower income tend to use public transportation, while those with a higher income prefer private vehicles. Similarly, individuals with higher incomes are more likely to frequent shopping, cultural entertainment, and tourist attraction POIs. By analysing the primary places of expenditure during travels at POIs, cost attributes can be deduced.

Safety features can be ascertained by noting an individual's travel habits. Those with a lower sense of security tend not to travel to remote locations or at late hours, reducing their likelihood of visiting remote public transportation facilities and entertainment categories. These individuals often visit healthcare and fitness POIs to enhance their personal safety. Additionally, their POI groups tend to be more simple and less diverse.

Lastly, personality traits can also be associated with POI visitation patterns. Extroverted individuals are more likely to frequent lively places like entertainment venues, shopping centres, and tourist attractions, whereas introverted individuals prefer quieter, independent environments and therefore visit them less frequently. However, introverts may have higher demands for services and sports activities.

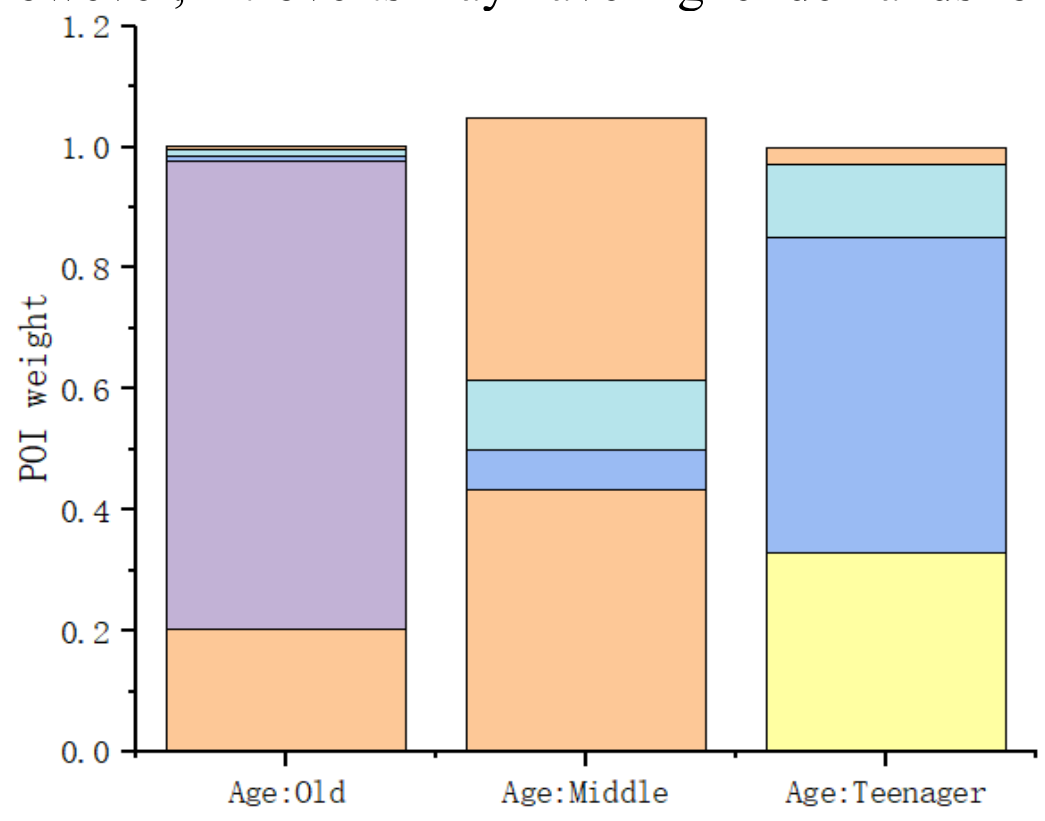

(a) Age attribute

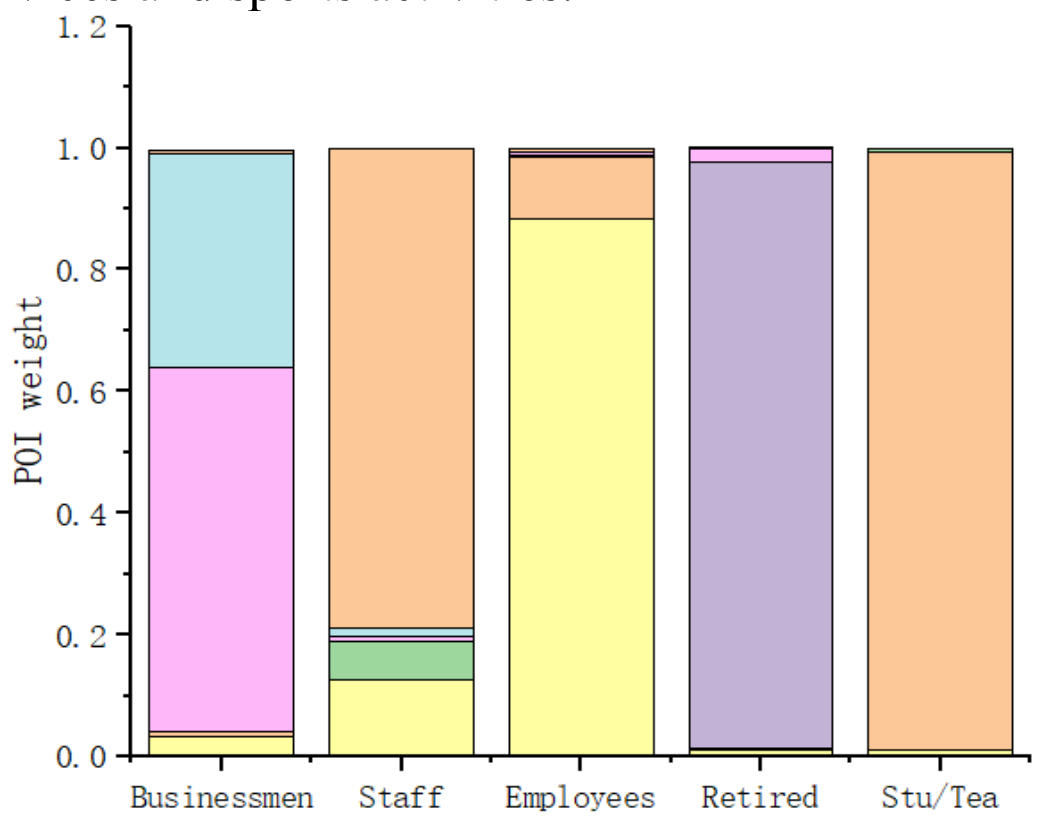

(b) Occupational attribute

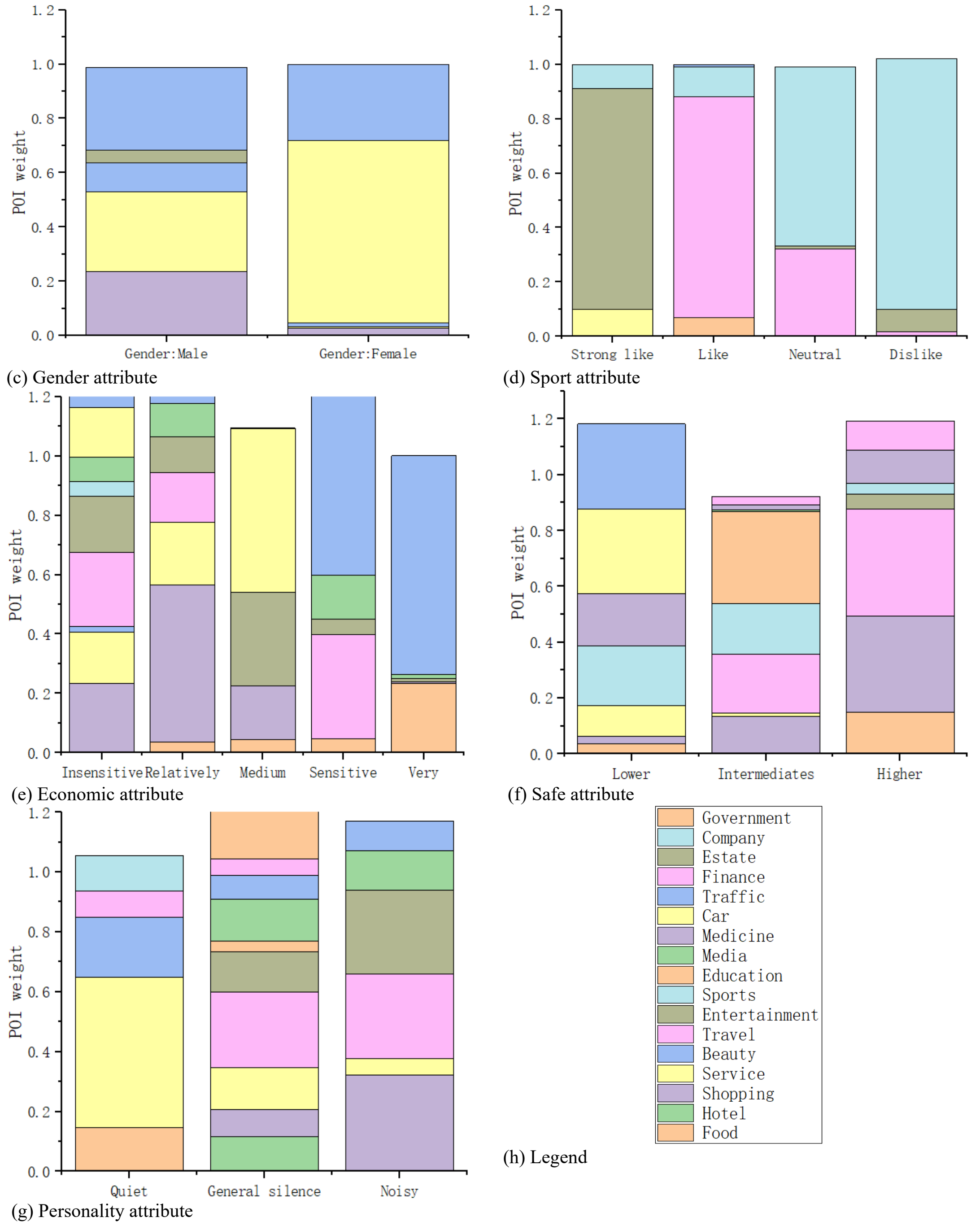

Fig. 10 LDA feature mining results.

In the LDA classification calculations, the attribute values for different individuals are unique. Therefore,

the optimal classification results are determined through a consistency score. After multiple experimental computations, a dataset with the highest consistency score was obtained. As displayed in Fig. 10, among the seven feature attributes - age, occupation, gender, sports, economy, security, and personality - the impact of each POI on the passenger varies.

While the weights of feature attributes across different passengers are not completely identical, a discernible digital feature representation is still apparent from the data results. For instance, the medical feature attributes carry a relatively higher weight for the elderly, retirees, and women, with retirees exhibiting the highest frequency of travel to these POI types. Men, corporate white-collar workers, and introverted individuals show a higher interest in vehicle-related POIs. Businesspeople concentrate on finance-associated POIs. Individuals with a low exercise frequency tend to gravitate towards recreational activities, while those with a high exercise frequency show a greater inclination towards sports-related POIs. Most of the trips made by students and educators are linked to educational institutions.

By integrating POI to deeply analyse passengers' travel trajectories and behaviours, it becomes evident that individual passenger attributes distinctly reflect in their respective travel actions. Thus, by enhancing feature extraction algorithms using POI, we can effectively and intuitively discern the relationship between different passenger behaviours and their travel characteristics, thereby laying a foundation for traffic optimization.

## 5. Conclusions

Bus travel offers a multitude of benefits, such as increasing urban traffic capacity, alleviating traffic congestion, saving energy, and decreasing urban emissions. Despite initial research primarily focusing on optimizing group travel through network planning, bus scheduling, and efficient bus resource allocation, there has been a recent shift towards personalization in bus transit services. This shift, spurred by the advent of big data and the rise of the Internet of Things, has necessitated new methodologies to understand and cater to individual travel needs better.

In response, we introduce a Point of Interest (POI) based passenger travel feature analysis method that allows a deeper understanding of passengers' attribute characteristics, enables the creation of comprehensive residential travel maps, and analyses travel patterns for future predictions. To fulfil the unique demands of feature mining in bus travel, we have designed two new methodologies: the P-KMEANS trajectory mining method and the P-LDA feature mining method. The model exploits POI seeds for pre-processing and generating distinctions within the algorithm's input parameters, simplifying the algorithm parameter selection process and bridging disparities in topic classification. This leads to a more efficient and optimized feature mining procedure.

As per the design, the P-KMEANS method can generate high-quality travel cluster results while the P-LDA method significantly boosts the accuracy of feature mining, providing weight feature indicators that closely mirror real-world conditions. In a practical setting, we first apply the P-KMEANS algorithm to group POIs within a specific range in the Yan'an region. Experimental results demonstrate the superiority of P-KMEANS over traditional KMEANS, using metrics such as the Silhouette Score Index, Calinski Harabasz Index, and Davies Bouldin Score. We then utilize the P-LDA model to incorporate various feature attributes, enabling the computation of passengers' feature attribute probabilities and creating a bus travel feature attribute model. Our experiments suggest that the results of this feature mining method outperform other algorithms on metrics like Recall, Precision, and Mean Absolute Error (MAE).

In conclusion, the public transport travel mining method, based on POI seeds presented in this research, permits the exploration of deep-level passenger attributes and the extraction of potential travel needs. This methodology facilitates a shift from individual to group travel and a transition from micro to macro-optimization of travel services.

Despite the validated algorithm performances, this work has limitations. The optimization strategy in both P-KMEANS and P-LDA models requires pre-training and pre-discrimination of the original model parameters.

While these modifications can enhance algorithmic performance, they simultaneously increase the time and space complexity of the algorithm, which can become burdensome when processing massive trajectory data, often leading to high costs. Hence, our future work will contemplate incorporating distributed computing methods involving data partitioning and blocking. This approach will strive to transition feature mining from a big data context to the utilization of distributed small data models, thus reducing both the time and space complexity associated with mining operations.